\ifdefined\pdftexversion
  \pdfoutput=1
\fi
\documentclass{mc_nankai}

\usepackage[T1]{fontenc}
\usepackage[numbers,square]{natbib}

\usepackage{amsmath,amsfonts,bm}

\def\eqref#1{equation~\ref{#1}}

\def\1{\bm{1}}

\DeclareMathAlphabet{\mathsfit}{\encodingdefault}{\sfdefault}{m}{sl}
\SetMathAlphabet{\mathsfit}{bold}{\encodingdefault}{\sfdefault}{bx}{n}

\usepackage{array}
\usepackage{wrapfig}
\usepackage{needspace}
\usepackage{url}
\usepackage{algorithm}
\usepackage{algpseudocode}
\usepackage{float}
\usepackage{colortbl}
\usepackage{makecell}
\usepackage{arydshln}
\usepackage{tabularx}
\usepackage{etoc}
\usepackage{amssymb}
\definecolor{dsOursRow}{HTML}{EAF2FA}
\title{DeltaStreamer: Towards Human-Like Agentic Streaming Video Understanding}
\hypersetup{pdftitle={DeltaStreamer: Towards Human-Like Agentic Streaming Video Understanding}}
\author[1]{Boyuan~Sun}
\author[2]{Deshui~Miao}
\author[2]{Wenzhao~Gao}
\author[2]{Shaoyong~Jia}
\author[2]{Shaohui~Jiao}
\author[3]{Chi~Liu}
\author[3]{Yan~Shu}
\author[3]{Derek~Li}
\author[3]{Bryan~Dai}
\author[4]{Shengsheng~Qian}
\author[1]{Qibin~Hou}

\affiliation[1]{VCIP, Nankai University}
\affiliation[2]{ByteDance Inc.\qquad\affiliationformat[3]{IQuest Research}}
\affiliation[4]{Institute of Automation, Chinese Academy of Sciences}

\hypersetup{pdfauthor={Boyuan Sun, Deshui Miao, Wenzhao Gao, Shaoyong Jia, Shaohui Jiao, Chi Liu, Yan Shu, Derek Li, Bryan Dai, Shengsheng Qian, Qibin Hou}}

\mclink[Project page]{\protect\url{https://bbbbchan.github.io/DeltaStreamer/}}
\abstract{We propose DeltaStreamer, a novel agentic framework for streaming video understanding inspired by how humans observe, remember, and investigate. Streaming video understanding requires a system to retain evidence before knowing which details future queries will need. Existing approaches often obscure event transitions and omit relevant details when compressing video history, and cannot recover missing visual evidence. DeltaStreamer addresses these limitations by combining change-aware memory with query-driven planning and targeted evidence gathering with replay. Before a query arrives, it continuously observes the stream and explicitly records visual changes in structured memory, preserving how events unfold rather than only what is present. When a query arrives, a query-driven planner identifies the required evidence and guides memory retrieval and targeted replay. By revisiting historical frames, the agent can recover visual details omitted during observation and ground its answer in evidence beyond the retained descriptions. This connects continuous observation with active investigation tailored to each question. Experiments on OVO-Bench and StreamingBench demonstrate improvements over representative streaming video understanding methods, with overall scores of 66.07\% and 59.53\%, respectively.}

\addto\captionsenglish{%
}
\begin{document}
\begingroup
\maketitle
\par
\endgroup

\section{Introduction}

Streaming video is prevalent in real-world applications such as live surveillance~\citep{feng2026monitor}, robotics~\citep{brohan2022rt,sun2025depth}, and autonomous systems~\citep{wang2023we,sun2023corrmatch}, where visual input arrives continuously rather than as a complete clip. Unlike offline video, where the entire sequence is available for analysis, streaming video arrives frame by frame. The system can access only what has already been observed and must operate continuously without seeing future frames~\citep{huang2025online}. Questions may also arrive after the relevant events~\citep{zhao2026omnipro}. The system must therefore build and maintain memory without knowing which details will later be needed.

Existing offline video models~\citep{zhao2025llava,sun2026see} struggle to adapt to the streaming setting, which requires continuous perception, incremental memory updates, and on-demand responses. Recent work has begun to address these challenges using token compression~\citep{chatterjee2025memory},
KV-cache reuse~\citep{kim2026infinipot, chen2026streamkv},
and multi-turn dialogue~\citep{xia2026streaming, wang2026streambridge}.
Despite the progress, these directions share two limitations in streaming scenes. First, their records primarily capture what is present while leaving what is changing implicit. Consecutive frames often contain similar scene content~\citep{zhao2026facial}, so repeated representations can consume the available budget and obscure the transitions that distinguish events. Second, once a frame has passed, the evidence is limited to what is retained from the visual signals~\citep{zhang2025memory}. 
Without a mechanism to revisit the original frame, details omitted during observation cannot be recovered when a later question makes them relevant. 
Both limitations reflect that existing paradigms focus on managing the token budget, with insufficient attention to recording changes and recovering missing evidence. This motivates us to investigate a more natural and intuitive way for streaming video understanding.

We argue that the system should watch frame by frame, just as a human does, continuously observing, building and updating a structured memory. As frames arrive sequentially~\citep{wang2026think,guan2026video} and queries arrive independently, the system needs to process visual information and update memory asynchronously~\citep{liu2026thinking,lin2026speak}. Before a query arrives, it should retain broad visual evidence without knowing which details will later matter. When a question arrives, it should actively probe its memory, assess whether the retrieved evidence is sufficient, and revisit historical frames to recover missing pixel-level details for a well-grounded answer. 
We should view streaming video understanding as an active information-gathering~\citep{zheng2026garde,wei2026see} problem rather than a passive encoding task, calling for an agent that extends a VLM with planning and tool use~\citep{wang2026streammeco}. Recent efforts such as StreamAgent~\citep{yang2025streamagent} and EventMemAgent~\citep{wen2026eventmemagent} have explored this agentic direction, but they share the same limitation that stored representations may omit evidence needed for later queries.

In this paper, we build DeltaStreamer, a human-like agentic streaming video understanding system that continuously builds a hierarchical textual memory from incoming frames and actively gathers evidence in response to user queries, just as human viewers follow a video as it unfolds, remember what happened, and revisit earlier footage when memory falls short. Considering that consecutive frames may yield similar scene descriptions even when important changes occur, the system compares adjacent frames during observation and explicitly records their visual changes as \textbf{deltas} alongside scene descriptions. These observations are organized into a hierarchical textual memory, with source links connecting summaries to their supporting observations and video frames. When a query arrives, a query-driven planner identifies the required evidence and whether it should be sought in past, current, or incoming frames. The resulting plan remains available throughout subsequent reasoning, guiding memory retrieval, continued observation, and tool use. When retrieved descriptions lack a required visual detail, the agent can follow the source links to locate relevant historical frames and replay them to recover pixel-level evidence. Unlike previous works that mostly focus on compressing and retaining video history, we couple explicit change recording with persistent query-driven planning and targeted replay. This allows the agent system to reason over event transitions and recover query-relevant visual details omitted from memory. We evaluate DeltaStreamer on OVO-Bench~\citep{niu2025ovo} and StreamingBench~\citep{lin2026streamingbench}, where it outperforms representative methods from existing streaming video understanding paradigms.


\begin{figure}[t]
\centering
\setlength{\abovecaptionskip}{2pt}
\includegraphics[width=0.96\textwidth]{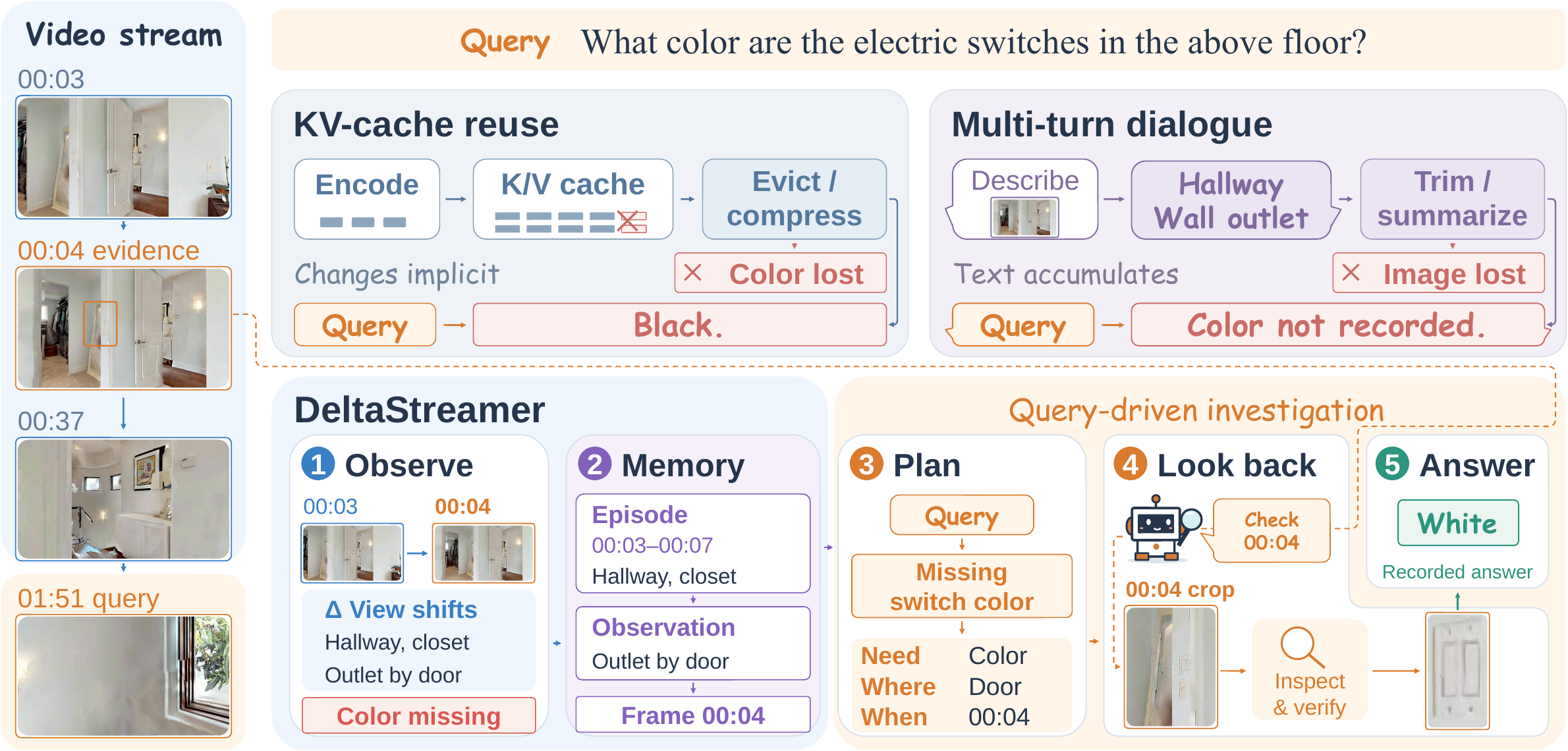}
\caption{Comparison of streaming video understanding paradigms. Compression or truncation in KV-cache reuse and multi-turn dialogue can obscure transitions and discard details needed by later queries. DeltaStreamer records explicit changes in structured memory and uses query-driven planning and targeted replay to recover missing visual evidence.
}
\vspace{-10pt}
\label{fig:paradign}
\end{figure}

In summary, our contributions are:
\begin{itemize}
    \item We propose DeltaStreamer, an agentic streaming video understanding system inspired by human observation and active investigation. It continuously builds structured memory and, after query arrival, uses planning and tool use to gather evidence until it can answer.
    \item We design a delta-augmented hierarchical textual memory that explicitly records inter-frame changes alongside scene descriptions. A three-layer DAG with bidirectional cross-layer pointers and an entity index makes these records traceable across temporal scales.
    \item We introduce a query-driven planner that performs structured analysis of the query's needs and produces a persistent plan to guide memory retrieval, continued observation, and tool use. Targeted replay recovers pixel-level visual details missing from textual memory.
\end{itemize}

\section{Related Work}
\label{sec:related_work}

\textbf{Streaming Video Representations and Memory.} Streaming video understanding requires continuous processing and selective retention of historical evidence. Existing approaches support these requirements through token compression~\citep{li2025videoscan, sun2025llava,wang2026accelerating,jin2025streamingassistant}, KV memory~\citep{tao2025infinitevl,agarwal2026going, sitong2026vinci2,zhang2026weavetime}, and multi-turn dialogue~\citep{yang2025svbench,xiong2025streaming,chen2024videollm}. TimeChat-Online~\citep{yao2025timechat} removes redundant tokens using inter-frame differences, ReKV~\citep{di2025streaming} retrieves historical KV states, and StreamMem~\citep{yang2025streammem} maintains compressed, query-agnostic KV memory. Streamo~\citep{xia2026streaming} develops instruction tuning for streaming interaction. Other approaches explicitly organize history for later queries. StreamForest~\citep{zeng2026streamforest} constructs an event-level memory forest, WAT~\citep{han2026wat} separates query-independent watching from query-triggered thinking, and Vista~\citep{lu2026vista} recalls full-resolution historical evidence from compressed scene representations. Together, these methods establish mechanisms for retaining and accessing video history. Some works also begin to enhance understanding quality from an omni-modal perspective~\citep{chen2025live, wang2025omnimmi,zhao2025humanomni,yang2025humanomniv2}. DeltaStreamer focuses on making semantic changes explicit and traceable. It directly compares adjacent sampled frames and stores the resulting deltas alongside scene descriptions in a hierarchical memory with source links and an entity index, connecting recorded transitions to supporting visual evidence.

\textbf{Agentic Evidence Gathering.} Streaming agents~\citep{zhao2026realtime, ko2026svmemagent, zhang2026streamscout} extend memory retrieval with planning and tool use. StreamAgent~\citep{yang2025streamagent} explores anticipatory planning for online interaction. EventMemAgent~\citep{wen2026eventmemagent} combines event memory, visual anchors, and inter-event change logs with learned memory search and visual tool use. OASIS~\citep{liang2026oasis} uses an adaptive coarse-to-fine process that generates retrieval intents and retrieves event memory with keyframes. Different from them, DeltaStreamer couples direct inter-frame change recording with a structured plan specifying the query target, temporal anchors, and required evidence. The plan remains available to guide subsequent retrieval and tool rounds, including targeted historical-frame replay when stored observations lack decisive details. 

\section{DeltaStreamer}
\label{sec:method}

\subsection{Problem Formulation and Overview}

\begin{figure}[t]
\centering
\setlength{\abovecaptionskip}{2pt}
\includegraphics[width=0.99\textwidth]{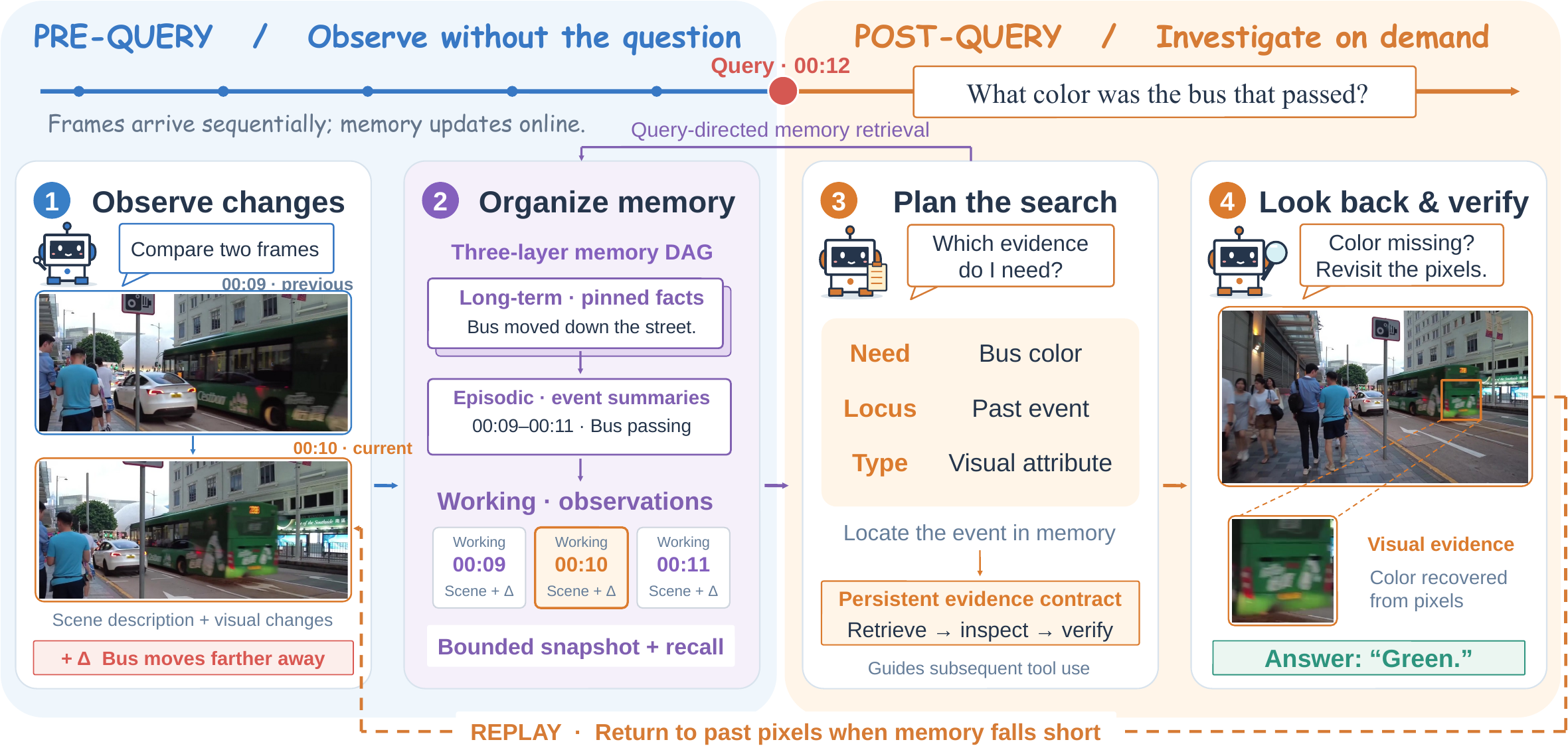}
\caption{Overview of DeltaStreamer. Before the query, adjacent-frame comparison produces delta-augmented observations stored in hierarchical memory. After the query, the planner guides evidence retrieval and tool use, while targeted replay recovers visual details for verification.}
\label{fig:overview}
\vspace{-12pt}
\end{figure}

Given a streaming video $\mathcal{V}=\{I_0,\ldots,I_{T-1}\}$ contain $T$ frames, where $I_t$ is the frame at index $t$. A question $q$ arrives at index $t_q$, which need not coincide with the time when sufficient evidence becomes available. Before $t_q$, the system observes the stream without access to $q$. After query arrival, it may answer using the evidence already available or continue observing incoming frames until it can respond. At any current index $t\geq t_q$, the accessible video is $\mathcal{V}_{\leq t}=\{I_0,\ldots,I_t\}$, including both pre-query frames and any frames received since query arrival, while frames beyond $t$ remain inaccessible. To address this setting, DeltaStreamer maintains textual memory as a hierarchical graph $\mathcal{G}_t$ through agentic tool use. During pre-query observation, it stores structured observations $O_t=(D_t,\Delta_t)$, where $D_t$ describes the current scene and $\Delta_t$ records changes between adjacent frames. At query arrival, a planner generates a plan $P$ to guide memory retrieval, continued observation, and targeted replay until the agent answers. The agent can inspect the current scene, continue observing, or use a targeted replay mechanism to recover missing visual details. Fig.~\ref{fig:overview} provides an overview of DeltaStreamer, including delta-augmented observation, traceable memory, query-driven planning, and replay.


\subsection{Delta-Augmented Observation}
\label{sec:delta}

Before query arrival, a vision-language model (VLM) jointly describes the current scene and its changes from the preceding sampled frame. The delta is a textual description of these changes, such as an object appearing or changing state.
\begin{equation}
O_t=(D_t,\Delta_t)
=\text{VLM}(I_{t-1},I_t,M_{t-1}^{\text{snapshot}}).
\label{eq:observation}
\end{equation}
Here, $t$ indexes sampled frames, $I_{t-1}$ and $I_t$ are the preceding and current frames, and $M_{t-1}^{\text{snapshot}}$ is the memory text available before processing $I_t$. The observation $O_t$ pairs a structured scene description $D_t$ (description, actors, objects, actions, OCR, and scene) with textual changes $\Delta_t$. At $t=0$, the model receives only $I_0$ and emits an initial-frame marker.


To identify repetitive descriptions, we measure the textual novelty of each nonempty observation against recent history,
\begin{equation}
s(O_t)=\frac{\sum_f w_f\left[1-\max_i\operatorname{Jaccard}_f\!\left(O_t,O_i\right)\right]}{\sum_f w_f}.
\label{eq:novelty}
\end{equation}
Here, $s(O_t)$ measures the textual novelty of observation $O_t$ at sampled frame $t$. The index $f$ ranges over fields with nonempty token sets in $O_t$, and $w_f$ is the positive weight assigned to field $f$. The index $i$ ranges over recent prior observations $O_i$. $\operatorname{Jaccard}_f(O_t,O_i)$ measures the intersection over union of the lowercase token sets extracted from field $f$ in the two observations. For each field, we subtract its highest historical similarity from one and then average these novelty scores using the field weights. Higher scores indicate less repetitive content. Novelty and repeated-delta checks assign salience weights for retrieval ranking and change flags for episodic aggregation. Low-salience records remain stored. The field weights and scoring procedure are specified in Appendix~\ref{app:observation}.

\subsection{Traceable Hierarchical Memory}
\label{sec:memory}

We formulate structured memory for long-running video streams as a hierarchical directed acyclic graph (DAG), with source pointers linking stored records to observations and video frames. Working memory stores working observations $\mathcal{W}$, episodic memory $ \mathcal{E} $ summarizes related observations, and long-term pins $ \mathcal{L} $ retain compact summaries linked to source episodes.

The memory DAG links $\mathcal{W}$, $\mathcal{E}$, and $\mathcal{L}$ through source identifiers and reverse pointers. After query arrival, retrieval and construction of the model's visible memory snapshot are expressed as:
\begin{equation}
\begin{aligned}
\mathcal{R}_t(q)&=\operatorname{Backtrack}(\mathcal{G}_t,q),\\
M_t^{\text{snapshot}}&=\operatorname{Render}\!\left(
\mathcal{L}_t,\mathcal{E}_t,\mathcal{W}_t^{\text{recent}},
\operatorname{Recall}(q,\mathcal{W}_t^{\text{middle}})\right).
\end{aligned}
\label{eq:memory}
\end{equation}
Here, $t$ indexes the current sampled frame, $q$ is the question, and $\mathcal{G}_t$ is the memory graph. $\operatorname{Backtrack}$ selects records by keyword overlap and salience, then follows source pointers to form evidence paths $\mathcal{R}_t(q)$. $\operatorname{Render}$ assembles snapshot $M_t^{\text{snapshot}}$ from long-term pins $\mathcal{L}_t$, episodes $\mathcal{E}_t$, recent working observations $\mathcal{W}_t^{\text{recent}}$, and question-relevant older observations selected by $\operatorname{Recall}$ from $\mathcal{W}_t^{\text{middle}}$.

The snapshot is the memory text supplied to each model call. Rendering trims older verbatim entries toward a text-budget target while retaining summaries and recent observations. Omitted records remain stored and retrievable. Before query arrival, recall can instead use recent observations as its retrieval cue. An entity index $\mathcal{J}_t$ maps actor and object names to their observation times. We describe memory construction and snapshot rendering in Appendix~\ref{app:memory}.

\subsection{Query-Driven Planner}
\label{sec:planner}

When a query arrives, a dedicated planner guides memory retrieval, continued observation, and tool use according to the question's information needs. It determines what evidence the question requires and whether to seek it in past, current, or incoming frames. Initial retrieval uses the question alone to provide context for the planning call.
\begin{equation}
\begin{aligned}
C_{t_q}&=\left(M_{t_q}^{\text{snapshot}},\mathcal{J}_{t_q},\mathcal{R}_{t_q}(q)\right),\\
P&=\text{VLM}(q,I_{t_q},C_{t_q}).
\end{aligned}
\label{eq:planner}
\end{equation}
Here, $q$ is the question, $t_q$ its arrival index, and $I_{t_q}$ the available frame. Context $C_{t_q}$ combines memory snapshot $M_{t_q}^{\text{snapshot}}$, entity index $\mathcal{J}_{t_q}$, and evidence paths $\mathcal{R}_{t_q}(q)$. The planning call $\text{VLM}$ produces plan $P$, specifying the target and temporal anchor, evidence type, temporal scope (past, current, future, or mixed), and search strategy. Auxiliary fields clarify viewpoint, available answer options, and whether the query concerns answerability or event progression (Appendix~\ref{app:planner}). Subsequent retrieval combines $q$ with the planned information need.

\begin{table*}[!t]
    \centering
    \normalsize
    \setlength{\abovecaptionskip}{2pt}
    \caption{The experimental results on the OVO-Bench benchmark. RT+BT Avg. is the average over the Real-Time Visual Perception and Backward Tracing groups.}
    \renewcommand{\arraystretch}{1.2}
    \setlength{\tabcolsep}{3pt}
    \resizebox{\linewidth}{!}{
        \begin{tabular}{l|ccccccc|cccc|c|cccc|c}
          \hline
          \multirow{2}{*}{Model} & \multicolumn{7}{c|}{Real-Time Visual Perception}  & \multicolumn{4}{c|}{Backward Tracing} & \multirow{2}{*}{\shortstack{\textbf{RT+BT}\\\textbf{Avg.}}} & \multicolumn{4}{c|}{Forward Active Responding} & \multirow{2}{*}{\textbf{Overall}} \\
          & OCR & ACR & ATR & STU & FPD & OJR & \textbf{Avg.} & EPM & ASI & HLD & \textbf{Avg.} & & REC & SSR & CRR & \textbf{Avg.} &  \\
          \hline
          \multicolumn{18}{c}{\textbf{Human}}\\
          \hline
          Human Agents & 93.96 & 92.57 & 94.83 & 92.70 & 91.09 & 94.02 & 93.20 & 92.59 & 93.02 & 91.37 & 92.33 & 92.77 & 95.48 & 89.67 & 93.56 & 92.90 & 92.81 \\
          \hline
          \multicolumn{18}{c}{\textbf{Proprietary MLLMs}}\\
          \hline
          Gemini 1.5 Pro & 85.91 & 66.97 & 79.31 & 58.43 & 63.37 & 61.96 & 69.32 & 58.59 & 76.35 & 52.64 & 62.54 & 65.93 & 35.53 & 74.24 & 61.67 & 57.15 & 63.00 \\
          Gemini 3 Flash & 86.58 & 68.81 & 79.31 & 58.99 & 70.30 & 70.65 & 72.44 & 58.92 & 76.35 & 52.15 & 62.47 & 67.46 & 35.53 & 74.24 & 61.67 & 57.15 & 64.02\\
          GPT-4o & 69.80 & 64.22 & 71.55 & 51.12 & 70.30 & 59.78 & 64.46 & 57.91 & 75.68 & 48.66 & 60.75 & 62.61 & 27.58 & 73.21 & 59.40 & 53.40 & 59.54 \\
          \hline
          \multicolumn{18}{c}{\textbf{Open-Source Offline MLLMs}}\\
          \hline
          Qwen2-VL-72B & 65.77 & 60.55 & 69.83 & 51.69 & 69.31 & 54.35 & 61.92 & 52.53 & 60.81 & 57.53 & 56.95 & 59.44 & 38.83 & 64.07 & 45.00 & 49.30 & 56.27 \\
          LLaVA-Video-7B & 69.13 & 58.72 & 68.83 & 49.44 & 74.26 & 59.78 & 63.52 & 56.23 & 57.43 & 7.53 & 40.40 & 51.96 & 34.10 & 69.95 & 60.42 & 54.82 & 52.91 \\
          LLaVA-OneVision-7B & 66.44 & 57.80 & 73.28 & 53.37 & 71.29 & 61.96 & 64.02 & 54.21 & 55.41 & 21.51 & 43.71 & 53.87 & 25.64 & 67.09 & 58.75 & 50.50 & 52.74 \\
          Qwen2-VL-7B & 60.40 & 50.46 & 56.03 & 47.19 & 66.34 & 55.43 & 55.98 & 47.81 & 35.48 & 56.08 & 46.46 & 51.22 & 31.66 & 65.82 & 48.75 & 48.74 & 50.39 \\
          InternVL-V2-8B & 67.11 & 60.55 & 63.79 & 46.07 & 68.32 & 56.52 & 60.39 & 48.15 & 57.43 & 24.73 & 43.44 & 51.92 & 26.50 & 59.14 & 54.14 & 46.60 & 50.15 \\
          LongVU-7B& 53.69 & 53.21 & 62.93 & 47.75 & 68.32 & 59.78 & 57.61 & 40.74 & 59.46 & 4.84 & 35.01 & 46.31 & 12.18 & 69.48 & 60.83 & 47.50 & 46.71 \\
          Qwen3-VL-8B & 75.17 & 58.72 & 72.41 & 57.30 & 70.30 & 59.24 & 65.52 & 56.57 & 69.59 & 12.37 & 46.18 & 55.85 & 47.13 & 67.57 & 52.50 & 55.73 & 55.81 \\
          \hline
          \multicolumn{18}{c}{\textbf{Limited Open-Source Online MLLMs}}\\
          \hline
          Stream-IT & 84.56 & 71.56 & 74.14 & 49.44 & 75.25 & 72.83 & 71.30& - & - & - & - & - & -& - & - & - & -\\
          HERMES & 85.20 & 64.20 & 71.60 & 53.40 & 74.30 & 65.20 & 69.00 & 48.50 & 62.20 & 37.60 & 49.40 & 59.20 & - & - & - &- & - \\
          VideoLLM-online-8B & 8.05 & 23.85 & 12.07 & 14.04 & 45.54 & 21.20 & 20.79 & 22.22 & 18.80 & 12.18 & 17.73 & 19.26 & - & - & - & - & - \\
          OASIS & 91.95 & 80.73 & 81.03 & 67.42 & 67.33 & 79.89 & 78.14 & 61.95 & 60.14 & 47.31 & 57.21 & 67.68 & - & - & - & - & - \\
          SimpleBaseline & 94.00 & 84.40 & 80.20 & 64.00 & 75.30 & 81.50 & 79.90 & 53.20 & 60.80 & 50.50 & 54.90 & 67.40 & - & - & - & - & - \\
          LatentStream & 85.90 & 61.50 & 73.30 & 46.60 & 78.20 & 65.20 & 68.50 & 50.80 & 60.80 & 68.30 & 60.00 & 64.20& - & - & - & - & - \\
          JoyAI-VL-Interaction & 72.50 & 61.50 & 75.00 & 59.00 & 78.20 & 64.10 & 68.40 & 62.00 & 70.90 & 12.90 & 48.60 & 58.50 & - & - & - & - & - \\
          Mage-VL & 94.63 & 83.49 & 79.31 & 74.16 & 70.30 & 77.17 & 79.84 & 50.17 & 61.49 & 32.80 & 48.15 & 64.00 & - & - & - & - & - \\
        FOLIO & 96.70 & 85.70 & 85.70 & 70.60 & 62.50 & 90.90 & 82.00 & 59.00 & 63.60 & 84.60 & 69.10 & 75.55 & - & - & - & - & - \\
          \hline
          \multicolumn{18}{c}{\textbf{Full Open-Source Online MLLMs  \& Video Agents }}\\
          \hline
          Flash-VStream-7B & 24.16 & 29.36 & 28.45 & 33.71 & 25.74 & 28.80 & 28.37 & 39.06 & 37.16 & 5.91 & 27.38 & 27.88 & 8.02 & 67.25 & 60.00 & 45.09 & 33.61 \\
          Dispider & 57.72 & 49.54 & 62.07 & 44.94 & 61.39 & 51.63 & 54.55 & 48.48 & 55.41 & 4.30 & 36.06 & 45.31 & 18.05 & 37.36 & 48.75 & 34.72 & 41.78 \\
          ReKV & 68.50 & 57.80 & 73.30 & 47.20 & 68.30 & 59.80 & 62.50 & 50.50 & 55.40 & 30.10 & 45.30 & 53.90 & 27.90 & 63.30 & 52.90 & 48.00 & 52.00\\
          StreamForest-7B & 68.46 & 53.21 & 71.55 & 47.75 & 65.35 & 60.87 & 61.20 & 58.92 & 64.86 & 32.26 & 52.02 & 56.61 & 32.81 & \underline{70.59} & 57.08 & 53.49 & 55.57 \\
          Streamo & 77.20 & 66.10 & \underline{76.70} & 45.50 & 66.30 & \underline{72.80} & 67.40 & 55.60 & 58.10 & 33.90 & 49.20 & 58.30 & 30.80 & 57.60 & \textbf{82.50} & \underline{57.00} & 57.90\\
          VST & \textbf{80.50} & 55.10 & 72.40 & 55.10 & \textbf{76.20} & 64.10 & 67.20 & 56.90 &64.90 & 48.40 & 56.70 & 61.95 & 33.00 & 66.90 & \underline{62.10} & 54.00 & 59.30\\
          ViSpeak & 75.17 & 58.72 & 71.55 & 51.12 & \underline{74.26} & 66.85 & 66.28 & \textbf{59.93} & 48.65 & \underline{63.98} & 57.52 & 61.90 & 33.81 & 68.52 & 60.42 & 54.25  & \underline{61.08}\\
          StreamAgent & 71.20 & 53.20 & 63.60 & 53.90 & 67.30 & 58.70 & 61.30 & 54.80 & 58.10 & 25.80 & 41.70 & 51.50 & \underline{35.90} & 48.40 & 52.00 & 45.40 & 49.40 \\
          EventMemAgent & 75.84 & \underline{69.72} & 73.28 & \underline{55.62} & 67.33 & 67.93 & \underline{68.29} & \underline{59.60} & \underline{70.95} & 43.55 & \underline{58.03} & \underline{63.16} & 33.67 & \textbf{72.02} & 62.08 & 55.92 & 60.75 \\
          \rowcolor{dsOursRow}
          \textbf{DeltaStreamer} & \underline{77.85} & \textbf{70.64} & \textbf{82.76} & \textbf{61.24} & \underline{74.26} & \textbf{73.68} & \textbf{72.23} & 54.21 & \textbf{79.05} & \textbf{69.89} & \textbf{67.72} & \textbf{69.98} & \textbf{54.87} & 59.46 & 60.42 & \textbf{58.25} & \textbf{66.07} \\
          \hline
        \end{tabular}
      }
    \label{tab:ovobench_result}
    \vspace{-10pt}
  \end{table*}
  
\textbf{Persistent evidence guidance.}
The plan $P$ is generated once and retained in subsequent prompts. Its accompanying evidence contract specifies where to seek evidence and when to use tools. The agent follows this guidance to retrieve memory, observe incoming frames, or replay historical frames, adapting its actions to updated memory and tool feedback. For details of the evidence contract and tool-use rules, please refer to Appendices~\ref{app:planner} and~\ref{app:tools}.

\subsection{Look-Back Verification and Replay}
\label{sec:replay}

To ground answers in visual evidence beyond stored descriptions, we combine memory consistency checks with targeted replay of historical frames. Replay has two entry points. The agent can request missing visual details, or a verifier can flag unresolved conflicts among retrieved records. Both routes inspect retained historical frames through $\operatorname{replay}(\tau)$, where $\tau$ is a target frame index available from the current index $t$, with $0\leq\tau\leq t$. 
Before a candidate answer is accepted, a rule-based verifier checks records retrieved using the planner's retrieval query. The verifier groups claims by subject and property and applies lightweight cross-layer consistency and temporal heuristics.
Unresolved conflicts with available historical frames become replay candidates.

For verifier-triggered candidates, heuristic priorities favor severe conflicts and frames closer to the question time. The scheduler selects
\begin{equation}
u^*=\operatorname*{arg\,max}_{u\in\mathcal{U}_t:\,\rho(u)>\eta}\rho(u),
\qquad \tau^*=\tau_{u^*},
\label{eq:replay-selection}
\end{equation}
Here, $t$ is the current frame index and $\mathcal{U}_t$ contains unresolved conflicts with available replay targets. For conflict $u$, $\rho(u)$ is its heuristic priority, $\eta$ the threshold, and $\tau_u$ its frame index. The highest-priority eligible conflict $u^*$ yields replay frame $\tau^*=\tau_{u^*}$. With no eligible candidate, the verifier requests no replay.
After inspecting the historical frame, the agent resumes investigation at the current frame, using additional visual evidence to generate an answer or re-evaluate. Appendix~\ref{app:verification} details the conflict rules, priorities, and candidate limit.

\begin{table*}[!t]
\centering
\normalsize
\setlength{\abovecaptionskip}{2pt}
\setlength{\abovecaptionskip}{2pt}
\caption{Evaluation results on StreamingBench~\citep{lin2026streamingbench}.
}
\label{tab:streaming_bench_full}
\renewcommand{\arraystretch}{1.5}
\setlength{\tabcolsep}{3pt}
\resizebox{\linewidth}{!}{
    \begin{tabular}{l|ccccccccccc|ccccc|ccccc|c}
\hline
\multirow{2}{*}{Method} & \multicolumn{11}{c|}{Real-Time Visual Understanding} & \multicolumn{5}{c|}{Omni-Source Understanding} & \multicolumn{5}{c|}{Contextual Understanding} & \multirow{2}{*}{\textbf{Overall}} \\
& OP & CR & CS & ATP & EU & TR & PR & SU & ACP & CT & \textbf{Avg.} & ER & SCU & SD & MA & \textbf{Avg.} & ACU & MCU & SQA & PO & \textbf{Avg.} & \\
\hline
\multicolumn{23}{c}{\textbf{Human}} \\
\hline
Human & 89.47 & 92.00 & 93.60 & 91.47 & 95.65 & 92.52 & 88.00 & 88.75 & 89.74 & 91.30 & 91.46 & 88.00 & 88.24 & 93.60 & 90.27 & 90.26 & 88.80 & 90.40 & 95.00 & 100.00 & 93.55 & 91.66 \\
\hline
\multicolumn{23}{c}{\textbf{Proprietary MLLMs}} \\
\hline
Gemini 1.5 Pro & 79.02& 80.47& 83.54& 79.67& 80.00& 84.74& 77.78& 64.23& 71.95& 48.70& 75.69& 46.80& 39.60& 74.90& 80.00& 60.22& 51.41& 40.73& 54.80& 45.10& 48.73& 67.07 \\
GPT-4o & 77.11& 80.47& 83.91& 76.47& 70.19& 83.80& 66.67& 62.19& 69.12& 49.22& 73.28& 41.20& 37.20& 43.60& 56.00& 44.50& 41.20& 38.40& 32.80& 56.86& 38.70& 60.15 \\
\hline
\multicolumn{23}{c}{\textbf{Open-Source Offline MLLMs}} \\
\hline
LLaVA-OV-7B & 80.38& 74.22& 76.03& 80.72& 72.67& 71.65& 67.59& 65.45& 65.72& 45.08& 71.12& 40.80& 37.20& 33.60& 44.80& 38.40& 35.60& 36.00& 27.27& 29.55& 32.74& 56.36 \\
Qwen2-VL-7B & 75.20& 82.81& 73.19& 77.45& 68.32& 71.03& 72.22& 61.19& 61.47& 46.11& 69.04& 41.20& 22.00& 32.80& 43.60& 34.90& 31.20& 26.00& 39.60& 22.73& 31.66& 54.14 \\
InternVL-V2-8B & 68.12& 60.94& 69.40& 77.12& 67.70& 62.93& 59.26& 53.25& 54.96& 56.48& 63.72& 37.60& 26.40& 37.20& 42.00& 35.80& 32.00& 31.20& 32.32& 40.91& 32.42& 51.40 \\
LongVA-7B & 70.03& 63.28& 61.20& 70.92& 62.73& 59.50& 61.11& 53.66& 54.67& 34.72& 59.96& 39.60& 32.40& 28.00& 41.60& 35.40& 32.80& 29.60& 30.30& 15.91& 29.95& 48.66 \\
VILA-1.5-8B & 53.68& 49.22& 70.98& 56.86& 53.42& 53.89& 54.63& 48.78& 50.14& 17.62& 52.32& 41.60& 26.40& 28.40& 36.00& 33.10& 26.80& 34.00& 23.23& 17.65& 27.35& 43.20 \\
Video-LLaMA2 & 55.86& 55.47& 57.41& 58.17& 52.80& 43.61& 39.81& 42.68& 45.61& 35.23& 49.52& 30.40& 32.40& 30.40& 36.00& 32.40& 24.80& 26.80& 18.67& 0.00& 21.93& 40.40 \\
Qwen3-VL-8B & 62.70& 68.00& 69.70& 53.30& 67.50& 65.10& 67.60& 48.00& 68.00& 40.90& 60.88& 36.80& 18.00& 32.80& 34.00& 30.40& 25.60& 23.60& 31.20& 32.40& 28.20& 46.84 \\
InternVL3-8B & 74.90& 82.00& 75.70& 61.20& 72.00& 67.60& 74.10& 66.70& 78.10& 34.20& 68.71& 40.40& 27.60& 38.80& 45.60& 38.10& 38.00& 26.00& 36.80& 31.20& 33.00& 53.97 \\
Gemma3-4B & 63.80& 69.50& 68.80& 54.40& 60.20& 65.10& 59.30& 40.70& 62.70& 21.80& 57.49& 28.80& 31.20& 30.40& 46.40& 34.20& 34.40& 31.20& 38.80& 12.40& 29.20& 46.03 \\
\hline
\multicolumn{23}{c}{\textbf{Limited Open-Source Online MLLMs}} \\
\hline
TimeChat-Online & 80.80& 79.70& 80.80& 83.30& 74.80& 78.80& 78.70& 64.20& 68.80& 58.00& 75.28& - & - & - & - & - & - & - & - & - & - & - \\
QueryStream & 82.11& 83.59& 78.23& 82.69& 75.47& 80.06& 79.63& 63.01& 67.90& 42.55& 74.04& - & - & - & - & - & - & - & - & - & - & - \\
LiveVLM & 79.84 & 79.69 & 84.86 & 80.72 & 67.08 & 70.09 & 74.07 & 66.26 & 67.99 & 42.49 & 72.36& - & - & - & - & - & - & - & - & - & - & - \\

\hline
\multicolumn{23}{c}{\textbf{Full Open-Source Online MLLMs \& Video Agents}} \\
\hline
Flash-VStream & 25.89& 43.57& 24.91& 23.87& 27.33& 13.08& 18.52& 25.20& 23.87& 48.70& 23.23& 25.91& 24.90& 25.60& 28.40& 26.00& 24.80& 25.20& 26.80& 1.96& 24.12& 24.04 \\
VideoLLM-Online & 39.07& 40.06& 34.49& 31.05& 45.54& 32.40& 31.48& 34.16& 42.49& 27.89& 35.99& 31.20& \underline{26.51}& 24.10& 32.00& 28.45& 24.19& 29.20& 30.80& 3.92& 26.55& 32.48 \\
Dispider & 74.92& 75.53& 74.10& 73.08& 74.44& 59.92& 76.14& 62.91& 62.16& 45.80& 67.63& 35.46& 25.26& 38.57& 43.34& 35.66& 39.62& 27.65& 34.80& 25.34& 33.61& 53.12 \\

STRIDE & 74.90& \textbf{78.90}& 76.70& 68.60& \textbf{77.00}& \underline{77.30}& 77.80& \underline{71.50}& \textbf{83.00}& 33.20& 72.45& \underline{39.60}& 22.40& \textbf{44.00}& \textbf{50.80}& \underline{39.20}& 34.00& \underline{35.20}& \underline{43.20}& \textbf{42.80}& \underline{38.80}& \underline{57.58}  \\
StreamAgent & \textbf{79.63}& \underline{78.31}& \underline{79.28}& \underline{75.87}& \underline{74.74}& 76.92& \textbf{82.94}& 66.31& \underline{73.69}& \textbf{55.40}& \underline{74.31}& 35.86& 26.26& 38.87& \underline{44.04}& 36.26& \underline{39.72}& 30.25& 39.60& \underline{28.90}& 34.62& 57.02 \\
\rowcolor{dsOursRow}
\textbf{DeltaStreamer} & \underline{77.80}& 72.70& \textbf{80.80}& \textbf{84.00}& 72.30& \textbf{81.90}& \underline{78.70}& \textbf{72.80}& 66.80& \underline{50.50}& \textbf{74.80}& \textbf{45.60}& \textbf{38.00}& \underline{40.00}& 44.00& \textbf{41.90}& \textbf{42.80}& \textbf{55.20}& \textbf{47.60}& 10.40& \textbf{39.00}& \textbf{59.53} \\
\hline
\end{tabular}
}
\end{table*}

\section{Experiments}
\label{sec:experiments}
\subsection{Experimental Details}

\textbf{Implementation Details.}
We use Qwen3.6-27B~\citep{qwen3.6-27b} as the shared backbone for observation, query-driven planning, and answer generation. Inference runs with vLLM~\citep{kwon2023efficient} in BF16 precision on 8x NVIDIA A100 GPUs. We process video frames at a base sampling rate of 1 fps using the backbone's image preprocessing pipeline.


\textbf{Benchmarks.}
We evaluate DeltaStreamer on OVO-Bench~\citep{niu2025ovo} and StreamingBench~\citep{lin2026streamingbench}. OVO-Bench tests real-time perception, recall of past events, and proactive responses, while StreamingBench covers real-time visual understanding, multimodal information, and contextual reasoning. Dataset statistics and task definitions are provided in Appendix~\ref{app:benchmarks}.


\subsection{Main Results}

\subsubsection{Compared Methods}
\label{sec:compared_methods}

We compare proprietary models, open-source offline models, and streaming systems. Proprietary models include Gemini 1.5 Pro~\citep{team2024gemini}, Gemini 3 Flash\footnote{\url{https://blog.google/products-and-platforms/products/gemini/gemini-3-flash/}}, and GPT-4o~\citep{openai2024gpt4o}.
Offline models include Qwen2-VL-7B/72B~\citep{Qwen2-VL}, Qwen3-VL-8B~\citep{bai2025qwen3}, LLaVA-Video-7B~\citep{zhang2024videoinstructiontuningsynthetic}, LLaVA-OneVision-7B (LLaVA-OV)~\citep{li2024llavaone}, InternVL2-8B~\citep{chen2024expanding}, InternVL3-8B~\citep{zhu2025internvl3}, LongVU-7B~\citep{shen2024longvu}, LongVA-7B~\citep{zhang2024long}, VILA-1.5-8B~\citep{lin2024vila}, VideoLLaMA2-7B~\citep{damonlpsg2024videollama2}, and Gemma3-4B~\citep{team2025gemma}.

Streaming systems include HERMES~\citep{zhang2026hermes}, VideoLLM-online~\citep{chen2024videollm}, OASIS~\citep{liang2026oasis}, SimpleBaseline~\citep{shen2026simple}, FOLIO~\citep{fan2026folio}, Stream-IT~\citep{wang2026streambridge}, JoyAI-VL-Interaction~\citep{yao2026joyai}, LatentStream~\citep{qu2026beyond}, Mage-VL~\citep{yang2026mage}, Flash-VStream~\citep{zhang2024flashvstream,zhang2025flash}, Dispider~\citep{qian2025dispider}, ReKV~\citep{di2025streaming}, StreamForest~\citep{zeng2026streamforest}, Streamo~\citep{xia2026streaming}, VST~\citep{guan2026videostreamingthinking}, ViSpeak~\citep{fu2025vispeak}, TimeChat-Online~\citep{yao2025timechat}, QueryStream~\citep{zhang2026querystream}, LiveVLM~\citep{ning2025livevlm}, STRIDE~\citep{kim2026stride}, StreamAgent~\citep{yang2025streamagent}, and EventMemAgent~\citep{wen2026eventmemagent}. 
Human results are provided by the respective benchmarks.

\begin{table*}[!t]
\centering
\small
\setlength{\abovecaptionskip}{2pt}
\caption{Ablation of delta-augmented observation. \textbf{2F-NoDelta} receives the same two-frame input as our method but does not explicitly generate or store a delta field. Comparing the two-frame variants tests explicit change recording at the same visual-input count.}
\label{tab:delta_ablation}
\renewcommand{\arraystretch}{1.0}
\setlength{\tabcolsep}{28.254pt}
\begin{tabular}{@{}lccc@{}}
\toprule
\textbf{ Setting }&\textbf{Input Frames} & \textbf{Delta Representation} & \textbf{OVO-Bench} \\
\midrule
1F-NoDelta & $I_t$ & No Delta & 58.87\\
1F-TextDelta & $I_t$ & Text-Memory comparison & 63.19\\
2F-NoDelta & $(I_{t-1},I_t)$ & No Delta & 61.12\\
2F-VisualDelta (Ours) & $(I_{t-1},I_t)$ & Visual comparison & \textbf{66.07}\\
\bottomrule
\end{tabular}
\end{table*}

\begin{table*}[!t]
\centering
\setlength{\abovecaptionskip}{2pt}
\caption{Ablation of memory components. All variants use identical observations, tools, planner, and answer decoding; only the component shown in each mini-table is changed.}
\label{tab:memory_ablation}
\footnotesize
\setlength{\abovecaptionskip}{2pt}
\renewcommand{\arraystretch}{1.16}
\captionsetup[subtable]{font=footnotesize,skip=4pt}
\begin{subtable}[t]{0.241\textwidth}
\centering
\setlength{\abovecaptionskip}{2pt}

\label{tab:memory_hierarchy_ablation}
\setlength{\tabcolsep}{0.05\linewidth}
\begin{tabular}{@{}p{0.64\linewidth}>{\centering\arraybackslash}p{0.26\linewidth}@{}}
\toprule
Flat Memory & 61.25 \\
Hierarchical & \textbf{66.07} \\
\bottomrule
\end{tabular}
\caption{Memory hierarchy}
\end{subtable}%
\hspace{0.012\textwidth}%
\begin{subtable}[t]{0.241\textwidth}
\centering
\setlength{\abovecaptionskip}{2pt}

\label{tab:memory_dag_ablation}
\setlength{\tabcolsep}{0.05\linewidth}
\begin{tabular}{@{}p{0.64\linewidth}>{\centering\arraybackslash}p{0.26\linewidth}@{}}
\toprule
w/o DAG & 62.73 \\
w/ DAG & \textbf{66.07} \\
\bottomrule
\end{tabular}
\caption{DAG organization}
\end{subtable}%
\hspace{0.012\textwidth}%
\begin{subtable}[t]{0.241\textwidth}
\centering

\label{tab:memory_entity_ablation}
\setlength{\tabcolsep}{0.05\linewidth}
\begin{tabular}{@{}p{0.64\linewidth}>{\centering\arraybackslash}p{0.26\linewidth}@{}}
\toprule
w/o Index & 63.40 \\
w/ Index & \textbf{66.07} \\
\bottomrule
\end{tabular}
\caption{Entity indexing}
\end{subtable}%
\hspace{0.012\textwidth}%
\begin{subtable}[t]{0.241\textwidth}
\centering

\label{tab:memory_episode_ablation}
\setlength{\tabcolsep}{0.05\linewidth}
\begin{tabular}{@{}p{0.64\linewidth}>{\centering\arraybackslash}p{0.26\linewidth}@{}}
\toprule
Fixed & 63.77 \\
BOCPD & \textbf{66.07} \\
\bottomrule
\end{tabular}
\caption{Episode Mode}
\end{subtable}
\end{table*}

\subsubsection{Benchmark Results}

\textbf{OVO-Bench.}
Tab.~\ref{tab:ovobench_result} reports an overall score of 66.07 for DeltaStreamer, a gain of $+5.32$ points over EventMemAgent. It also outperforms Gemini 3 Flash by $+2.05$ points, with higher scores on action-sequence identification, hallucination detection, and repeated-event counting. The method reaches 79.05 on action-sequence identification (ASI) and 69.89 on hallucination detection (HLD). Its improvements on these tasks align with our emphasis on distinguishing events and gathering evidence relevant to a question.

\textbf{StreamingBench.}
Tab.~\ref{tab:streaming_bench_full} reports an overall score of 59.53, exceeding STRIDE by $+1.95$ points and also outperforming StreamAgent. Contextual understanding improves over STRIDE, with scores of 55.20 on MCU and 47.60 on SQA. However, PO remains weak at 10.40, showing that improved evidence handling does not by itself solve proactive response timing. 
Together, these results demonstrate the benefits of change-aware memory and planned evidence gathering for temporal reasoning and contextual understanding.

\subsection{Ablation Studies}

\textbf{Effect of Explicit Visual Delta.}
To separate explicit change recording from additional visual input, we compare observation variants in Tab.~\ref{tab:delta_ablation}. With identical two-frame input,
\par
\Needspace{17\baselineskip}
\begingroup
\setlength{\intextsep}{4pt}
\setlength{\columnsep}{12pt}
\begin{wrapfigure}{r}{0.40\textwidth}
\centering
\includegraphics[width=\linewidth]{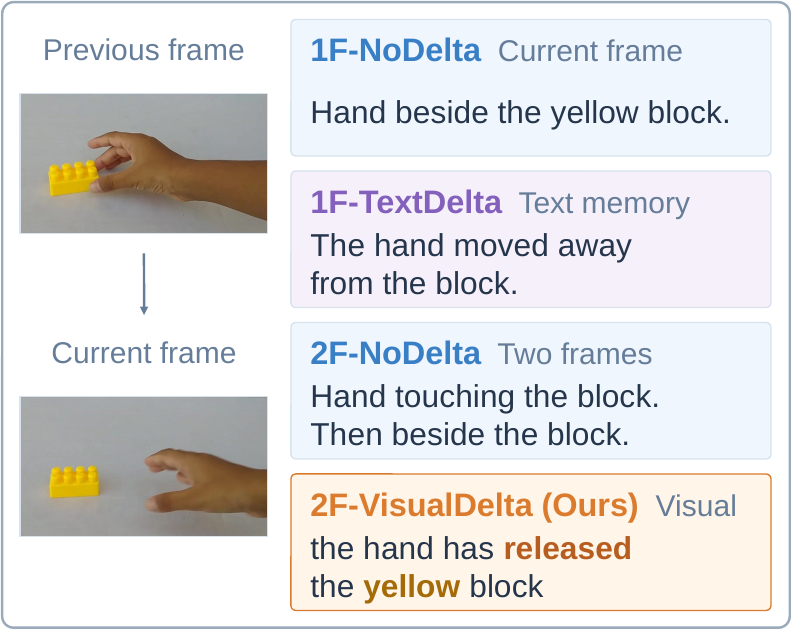}
\captionsetup{font=footnotesize,skip=4pt}
\caption{Illustrative delta comparison.}
\label{fig:delta_example}
\end{wrapfigure}
\noindent 2F-NoDelta scores 61.12 and 2F-VisualDelta reaches 66.07; visual comparison also outperforms 1F-TextDelta. These results support making temporal evidence explicit for later reasoning. Similar scene descriptions can leave event progression implicit, requiring reconstruction across separate records. Delta writes these transitions directly into memory, providing a retrieval cue for questions about how events unfold. Whereas 1F-TextDelta compares the current frame with textual memory, adjacent-frame comparison can use visual details omitted from earlier descriptions. The resulting observation retains both the current state and an explicit account of its changes, making the temporal relation directly available during investigation. We also demonstrate a comparison of these four contracts in Fig.~\ref{fig:delta_example}.
\par
\endgroup

\begin{table}[!t]
\centering
\setlength{\abovecaptionskip}{2pt}
\caption{Planner and component ablations on OVO-Bench. Left: cumulative planner settings, with persistence retaining the plan across post-query rounds. Right: combinations of delta (D), hierarchical memory (M), and planner (P).}
\label{tab:planner_ablation}
\captionsetup[subtable]{font=footnotesize,skip=3pt}
\begin{subtable}[t]{0.468\textwidth}
\vspace{0pt}
\setlength{\abovecaptionskip}{2pt}
\centering
\small
\renewcommand{\arraystretch}{1.28}
\setlength{\tabcolsep}{6pt}
\begin{tabular}{@{}>{\raggedright\arraybackslash}p{\dimexpr\linewidth-99pt\relax}>{\centering\arraybackslash}p{25pt}>{\centering\arraybackslash}p{50pt}@{}}
\toprule
\textbf{Cumulative setting} & \textbf{HLD} & \textbf{OVO-Bench} \\
\midrule
No planner & 45.28 & 59.30 \\
$+$ Plan generation & 65.83 & 63.85 \\
$+$ Persistence & 69.13 & 64.77 \\
$+$ POV & 68.07 & 65.14 \\
$+$ Evidence contract & \textbf{69.89} & \textbf{66.07} \\
\bottomrule
\end{tabular}
\par\vspace{21.2pt}
\caption{Planner ablation}
\label{tab:planner_progression}
\end{subtable}%
\hspace{0.012\textwidth}%
\begin{subtable}[t]{0.52\textwidth}
\vspace{-2.5pt}
\setlength{\abovecaptionskip}{2pt}
\centering
\includegraphics[width=\linewidth]{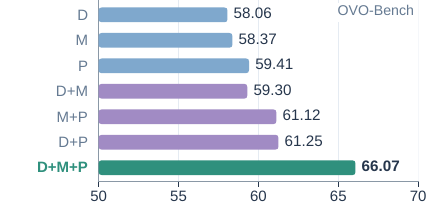}
\caption{Component combinations}
\label{tab:component_combinations}
\end{subtable}
\vspace{-10pt}
\end{table}

\begin{table*}[!t]
\centering
\setlength{\abovecaptionskip}{2pt}
\caption{Ablation of memory rendering and core visual tools. All variants keep the model, observation protocol, the same planner, and answer decoding fixed.}
\label{tab:render_tool_ablation}
\renewcommand{\arraystretch}{1.12}
\captionsetup[subtable]{font=footnotesize,skip=3pt}
\begin{subtable}[t]{0.44\textwidth}
\centering
\footnotesize
\setlength{\tabcolsep}{4pt}
\begin{tabular}{@{}>{\raggedright\arraybackslash}p{\dimexpr\linewidth-53pt\relax}>{\centering\arraybackslash}p{45pt}@{}}
\toprule
\textbf{Memory Rendering} & \textbf{OVO-Bench} \\
\midrule
Full History & 59.18  \\
Recent Tail Only & 60.04  \\
Pins + Episodes + Tail & 63.69  \\
Hierarchical + Middle Recall & \textbf{66.07}  \\
\bottomrule
\end{tabular}
\caption{Memory rendering strategy}
\label{tab:memory_rendering_ablation}
\end{subtable}%
\hspace{0.012\textwidth}%
\begin{subtable}[t]{0.548\textwidth}
\centering
\footnotesize
\setlength{\tabcolsep}{3pt}
\begin{tabular}{@{}>{\raggedright\arraybackslash}p{\dimexpr\linewidth-151pt\relax}>{\centering\arraybackslash}p{26pt}>{\centering\arraybackslash}p{33pt}>{\centering\arraybackslash}p{22pt}>{\centering\arraybackslash}p{46pt}@{}}
\toprule
\textbf{Configuration} & \textbf{Replay} & \textbf{Tracking} & \textbf{Zoom} &
\textbf{OVO-Bench} \\
\midrule
Full Tool Suite & \checkmark & \checkmark & \checkmark & \textbf{66.07}  \\
$-$ Replay &  & \checkmark & \checkmark & 63.28  \\
$-$ Tracking & \checkmark &  & \checkmark & 65.31  \\
$-$ Zoom & \checkmark & \checkmark &  & 64.16  \\
\bottomrule
\end{tabular}
\caption{Core visual tools}
\label{tab:tool_ablation}
\end{subtable}
\vspace{-10pt}
\end{table*}

\textbf{Memory Organization.}
Hierarchical memory reaches 66.07, compared with 61.25 for flat memory (Tab.~\ref{tab:memory_ablation}). The DAG and entity-index ablations also favor retaining both forms of access. As the stream grows, summaries provide an event-level overview, while source links and entity indexing help retrieve the underlying observations and timestamps when later queries require omitted details. History can thus be summarized while preserving a route to finer evidence. Online episode construction also outperforms fixed grouping, consistent with adapting memory to changes in the stream. This comparison covers the complete episode pipeline (Appendix~\ref{app:memory}).

\textbf{Query-Driven Planning.}
With cumulative planner additions in Tab.~\ref{tab:planner_ablation}(\subref{tab:planner_progression}), plan generation reaches 63.85 and persistence reaches 64.77. Planning makes the target, temporal anchor, and required evidence explicit, supplying retrieval cues beyond the query. Persistence keeps these requirements available as observations and tool results accumulate, tying later decisions to the original information need. The POV field specifies the relevant perspective, while the evidence contract guides retrieval, continued observation, and tool use. The substantial HLD improvement from plan generation highlights the importance of the plan itself for hallucination detection.

\textbf{Cross-Component Ablation.}
We examine the complementarity of delta (D), hierarchical memory (M), and planning (P) in Tab.~\ref{tab:planner_ablation}(\subref{tab:component_combinations}); configurations without M use flat memory. With M enabled, adding D yields $+0.93$ without P and $+4.95$ with P. This larger conditional benefit suggests that planning helps the agent use retained transitions. Richer memory is useful when the agent can identify which evidence a question requires, while planning benefits from explicit changes and traceable source records. This interaction is consistent with the full combination outperforming all single-component and pairwise settings.

\begin{figure}[t]
\centering
\setlength{\abovecaptionskip}{2pt}
\includegraphics[width=0.99\textwidth]{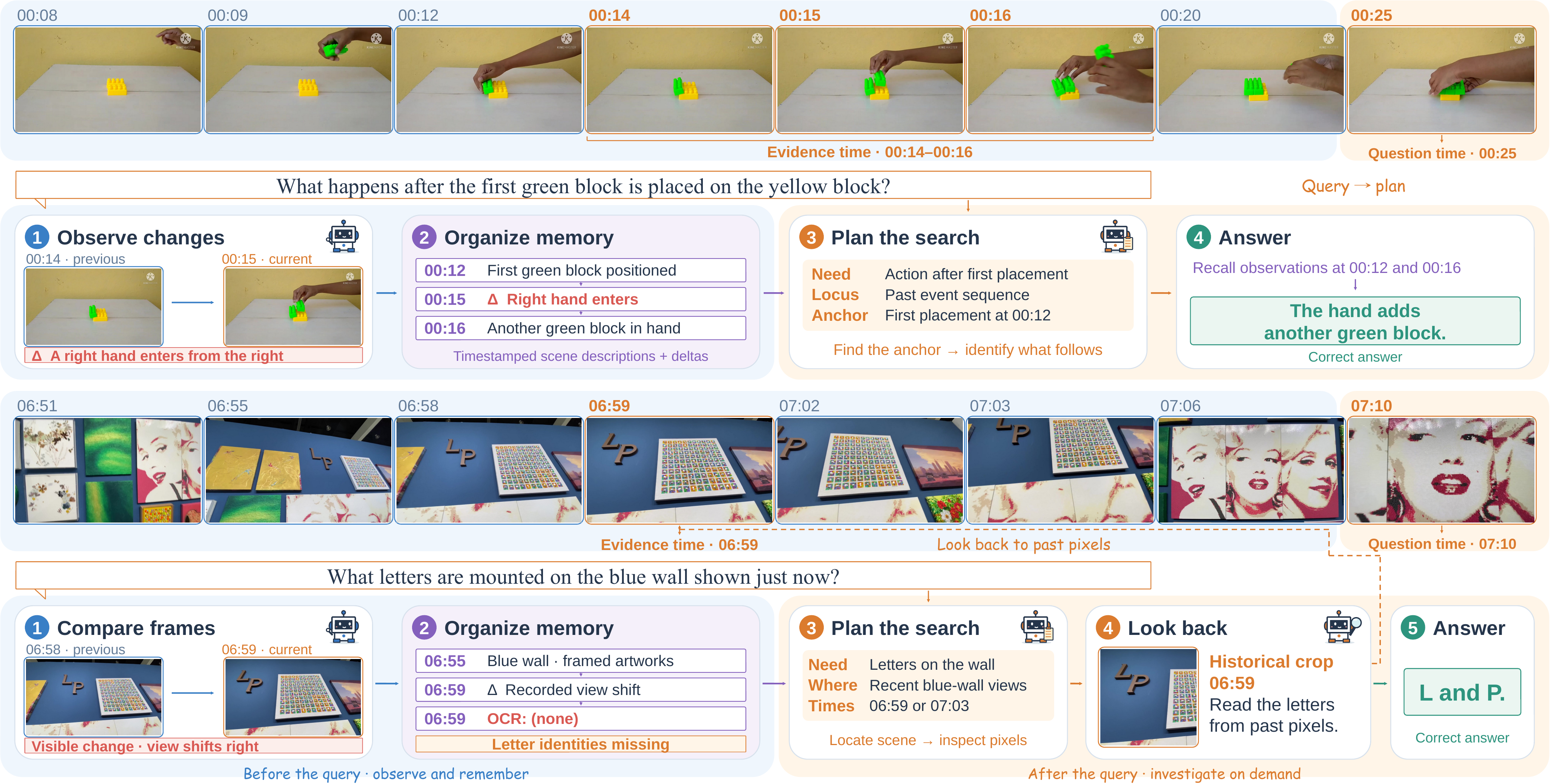}
\caption{\textbf{Qualitative examples of DeltaStreamer. }The upper case recalls the action following the first block placement. The lower case recovers the missing wall letters from a historical crop.}
\label{fig:qualitative}
\vspace{-5pt}
\end{figure}

\textbf{Memory Rendering and Tool Ablation.}
The rendering comparison examines how retained history enters the visible context (Tab.~\ref{tab:render_tool_ablation}). Adding middle recall to pins, episodes, and the recent tail raises accuracy from 63.69 to 66.07. Summaries preserve temporal coverage but cannot retain every detail a later question may require. Selective recall restores relevant older observations that are absent from the summaries and outside the recent window. The model can therefore use a compact temporal overview together with detailed source evidence selected for the question.

Removing replay lowers accuracy to 63.28, the largest drop among tool removals. Pre-query observation cannot anticipate every future information need. Middle recall recovers details already stored as text, while replay can supply visual evidence that was never encoded, explaining its value alongside structured memory. Zoom and tracking also help by exposing fine spatial details and supporting correspondence of the same target across frames. These capabilities address complementary evidence needs; each ablation removes one capability with the remaining tools available.

\subsection{Qualitative Analysis}
\label{sec:qualitative}

Fig.~\ref{fig:qualitative} illustrates the complementary roles of change-aware memory and query-driven evidence gathering. Explicit changes help represent event progression when successive frames contain similar scene content, while observations provide temporal anchors for later reasoning. However, textual memory can retain a scene without preserving every detail that a future question may require. The planner connects the query's information needs to relevant observations and guides historical visual inspection when stored descriptions are insufficient. This enables the agent to reason over retained evidence and recover omitted visual details on demand.

\section{Conclusions}

In this work, we introduce DeltaStreamer, a human-like agentic framework for streaming video understanding. It records explicit inter-frame changes in traceable hierarchical memory to preserve event transitions for later queries. A query-driven planner specifies evidence needs and guides memory retrieval and targeted replay when evidence is insufficient. Experiments on OVO-Bench and StreamingBench, together with ablations, demonstrate the benefits of combining change-aware memory with planned evidence gathering.
These findings support a streaming paradigm in which systems continuously observe and remember, then actively investigate when a question arrives.





\bibliography{references}

\begin{thebibliography}{84}
\providecommand{\natexlab}[1]{#1}
\providecommand{\url}[1]{\texttt{#1}}
\expandafter\ifx\csname urlstyle\endcsname\relax
  \providecommand{\doi}[1]{doi: #1}\else
  \providecommand{\doi}{doi: \begingroup \urlstyle{rm}\Url}\fi

\bibitem[Adams and MacKay(2007)]{adams2007bayesian}
Ryan~Prescott Adams and David~JC MacKay.
\newblock Bayesian online changepoint detection.
\newblock \emph{arXiv preprint arXiv:0710.3742}, 2007.

\bibitem[Agarwal et~al.(2026)Agarwal, Suri, Gwilliam, Kumar, and
  Shrivastava]{agarwal2026going}
Vatsal Agarwal, Saksham Suri, Matthew Gwilliam, Pulkit Kumar, and Abhinav
  Shrivastava.
\newblock Going down memory lane: Scaling tokens for video stream understanding
  with dynamic kv-cache memory.
\newblock \emph{arXiv preprint arXiv:2602.18434}, 2026.

\bibitem[Bai et~al.(2025)Bai, Cai, Chen, Chen, Chen, Cheng, Deng, Ding, Gao,
  Ge, et~al.]{bai2025qwen3}
Shuai Bai, Yuxuan Cai, Ruizhe Chen, Keqin Chen, Xionghui Chen, Zesen Cheng,
  Lianghao Deng, Wei Ding, Chang Gao, Chunjiang Ge, et~al.
\newblock Qwen3-vl technical report.
\newblock \emph{arXiv preprint arXiv:2511.21631}, 2025.

\bibitem[Brohan et~al.(2022)Brohan, Brown, Carbajal, Chebotar, Dabis, Finn,
  Gopalakrishnan, Hausman, Herzog, Hsu, et~al.]{brohan2022rt}
Anthony Brohan, Noah Brown, Justice Carbajal, Yevgen Chebotar, Joseph Dabis,
  Chelsea Finn, Keerthana Gopalakrishnan, Karol Hausman, Alex Herzog, Jasmine
  Hsu, et~al.
\newblock Rt-1: Robotics transformer for real-world control at scale.
\newblock \emph{arXiv preprint arXiv:2212.06817}, 2022.

\bibitem[Chatterjee et~al.(2025)Chatterjee, Remelli, Song, Tekin, Mittal,
  Bhatnagar, Camg{\~A}{\c{k}}z, Hampali, Sauser, Ma,
  et~al.]{chatterjee2025memory}
Dibyadip Chatterjee, Edoardo Remelli, Yale Song, Bugra Tekin, Abhay Mittal,
  Bharat Bhatnagar, Necati~Cihan Camg{\~A}{\c{k}}z, Shreyas Hampali, Eric
  Sauser, Shugao Ma, et~al.
\newblock Memory-efficient streaming videollms for real-time procedural video
  understanding.
\newblock \emph{arXiv preprint arXiv:2504.13915}, 2025.

\bibitem[Chen et~al.(2024{\natexlab{a}})Chen, Lv, Wu, Lin, Song, Gao, Liu, Gao,
  Mao, and Shou]{chen2024videollm}
Joya Chen, Zhaoyang Lv, Shiwei Wu, Kevin~Qinghong Lin, Chenan Song, Difei Gao,
  Jia-Wei Liu, Ziteng Gao, Dongxing Mao, and Mike~Zheng Shou.
\newblock Videollm-online: Online video large language model for streaming
  video.
\newblock In \emph{2024 IEEE/CVF Conference on Computer Vision and Pattern
  Recognition (CVPR)}, pages 18407--18418. IEEE, 2024{\natexlab{a}}.

\bibitem[Chen et~al.(2025)Chen, Zeng, Lin, Li, Ma, and Shou]{chen2025live}
Joya Chen, Ziyun Zeng, Yiqi Lin, Wei Li, Zejun Ma, and Mike~Zheng Shou.
\newblock Live: Learning video llm with streaming speech transcription at
  scale.
\newblock In \emph{2025 IEEE/CVF Conference on Computer Vision and Pattern
  Recognition (CVPR)}, pages 29083--29095. IEEE, 2025.

\bibitem[Chen et~al.(2026)Chen, Bai, Wang, Bai, Dai, and Lu]{chen2026streamkv}
Yilong Chen, Xiang Bai, Zhibin Wang, Chengyu Bai, Yuhan Dai, and Ming Lu.
\newblock Streamkv: Streaming video question-answering with segment-based kv
  cache retrieval and compression.
\newblock In \emph{Proceedings of the AAAI Conference on Artificial
  Intelligence}, volume~40, pages 3120--3128, 2026.

\bibitem[Chen et~al.(2024{\natexlab{b}})Chen, Wang, Cao, Liu, Gao, Cui, Zhu,
  Ye, Tian, Liu, et~al.]{chen2024expanding}
Zhe Chen, Weiyun Wang, Yue Cao, Yangzhou Liu, Zhangwei Gao, Erfei Cui, Jinguo
  Zhu, Shenglong Ye, Hao Tian, Zhaoyang Liu, et~al.
\newblock Expanding performance boundaries of open-source multimodal models
  with model, data, and test-time scaling.
\newblock \emph{arXiv preprint arXiv:2412.05271}, 2024{\natexlab{b}}.

\bibitem[Cheng et~al.(2024)Cheng, Leng, Zhang, Xin, Li, Chen, Zhu, Zhang, Luo,
  Zhao, and Bing]{damonlpsg2024videollama2}
Zesen Cheng, Sicong Leng, Hang Zhang, Yifei Xin, Xin Li, Guanzheng Chen,
  Yongxin Zhu, Wenqi Zhang, Ziyang Luo, Deli Zhao, and Lidong Bing.
\newblock Videollama 2: Advancing spatial-temporal modeling and audio
  understanding in video-llms.
\newblock \emph{arXiv preprint arXiv:2406.07476}, 2024.
\newblock URL \url{https://arxiv.org/abs/2406.07476}.

\bibitem[Di et~al.(2025)Di, Yu, Zhang, Li, Cheng, Li, He, Shu, and
  Jiang]{di2025streaming}
Shangzhe Di, Zhelun Yu, Guanghao Zhang, Haoyuan Li, Hao Cheng, Bolin Li,
  Wanggui He, Fangxun Shu, and Hao Jiang.
\newblock Streaming video question-answering with in-context video kv-cache
  retrieval.
\newblock In \emph{International Conference on Learning Representations},
  volume 2025, pages 42115--42127, 2025.

\bibitem[Fan et~al.(2026)Fan, Parikh, Ramachandran, Gobriel, Jain, Kannan, and
  Prasanna]{fan2026folio}
Haoyang Fan, Dhruv Parikh, Anvitha Ramachandran, Sameh Gobriel, Nilesh Jain,
  Rajgopal Kannan, and Viktor Prasanna.
\newblock Folio: Focused semantic memory for streaming video understanding.
\newblock \emph{arXiv preprint arXiv:2607.13298}, 2026.

\bibitem[Feng et~al.(2026)Feng, Liu, Zhang, Qin, et~al.]{feng2026monitor}
Yue Feng, Yingshi Liu, Jingrou Zhang, Jie Qin, et~al.
\newblock Monitor: Exploiting large language models with instruction for online
  video anomaly detection.
\newblock \emph{Advances in Neural Information Processing Systems},
  38:\penalty0 35851--35871, 2026.

\bibitem[Fu et~al.(2025)Fu, Yang, Li, Peng, Lin, Wei, Hu, Xie, and
  Zheng]{fu2025vispeak}
Shenghao Fu, Qize Yang, Yuan-Ming Li, Yi-Xing Peng, Kun-Yu Lin, Xihan Wei,
  Jian-Fang Hu, Xiaohua Xie, and Wei-Shi Zheng.
\newblock Vispeak: Visual instruction feedback in streaming videos.
\newblock In \emph{2025 IEEE/CVF International Conference on Computer Vision
  (ICCV)}, pages 21778--21788. IEEE, 2025.

\bibitem[Guan et~al.(2026{\natexlab{a}})Guan, Yin, Liang, Ju, Luo, Luan, Liu,
  and Bai]{guan2026video}
Yiran Guan, Liang Yin, Dingkang Liang, Jianzhong Ju, Zhenbo Luo, Jian Luan,
  Yuliang Liu, and Xiang Bai.
\newblock Video streaming thinking: Videollms can watch and think
  simultaneously.
\newblock \emph{arXiv preprint arXiv:2603.12262}, 2026{\natexlab{a}}.

\bibitem[Guan et~al.(2026{\natexlab{b}})Guan, Yin, Liang, Ju, Luo, Luan, Liu,
  and Bai]{guan2026videostreamingthinking}
Yiran Guan, Liang Yin, Dingkang Liang, Jianzhong Ju, Zhenbo Luo, Jian Luan,
  Yuliang Liu, and Xiang Bai.
\newblock Video streaming thinking: Videollms can watch and think
  simultaneously.
\newblock In \emph{European Conference on Computer Vision (ECCV)},
  2026{\natexlab{b}}.

\bibitem[Han et~al.(2026)Han, Sun, Xu, Tang, Lei, Zhang, Sun, He, and
  Sun]{han2026wat}
Zifan Han, Hongbo Sun, Jinglin Xu, Canhui Tang, Yulong Lei, Xuchong Zhang,
  Hongbin Sun, Zhongjiang He, and Hao Sun.
\newblock Wat: Online video understanding needs watching before thinking.
\newblock \emph{arXiv preprint arXiv:2603.13412}, 2026.

\bibitem[Huang et~al.(2025)Huang, Li, Li, Wang, Zeng, Liang, Wu, Chen, Li, and
  Wang]{huang2025online}
Zhenpeng Huang, Xinhao Li, Jiaqi Li, Jing Wang, Xiangyu Zeng, Cheng Liang, Tao
  Wu, Xi~Chen, Liang Li, and Limin Wang.
\newblock Online video understanding: Ovbench and videochat-online.
\newblock In \emph{2025 IEEE/CVF Conference on Computer Vision and Pattern
  Recognition (CVPR)}, pages 3328--3338. IEEE, 2025.

\bibitem[Jin et~al.(2025)Jin, Yu, Yu, Liu, Liu, Tao, Pei, Wang, Dang, Liu,
  et~al.]{jin2025streamingassistant}
Xinqi Jin, Hanxun Yu, Bohan Yu, Kebin Liu, Jian Liu, Keda Tao, Yixuan Pei, Huan
  Wang, Fan Dang, Jiangchuan Liu, et~al.
\newblock Streamingassistant: Efficient visual token pruning for accelerating
  online video understanding.
\newblock \emph{arXiv preprint arXiv:2512.12560}, 2025.

\bibitem[Kim et~al.(2026{\natexlab{a}})Kim, Lee, M.~Rehg, Kim, and
  Man~Ro]{kim2026stride}
Junho Kim, Hosu Lee, James M.~Rehg, Minsu Kim, and Yong Man~Ro.
\newblock Stride: When to speak meets sequence denoising for streaming video
  understanding.
\newblock In \emph{European Conference on Computer Vision}, pages 38--56.
  Springer, 2026{\natexlab{a}}.

\bibitem[Kim et~al.(2026{\natexlab{b}})Kim, Shim, Choi, and
  Chang]{kim2026infinipot}
Minsoo Kim, Kyuhong Shim, Jungwook Choi, and Simyung Chang.
\newblock Infinipot-v: Memory-constrained kv cache compression for streaming
  video understanding.
\newblock \emph{Advances in Neural Information Processing Systems},
  38:\penalty0 138983--139013, 2026{\natexlab{b}}.

\bibitem[Ko et~al.(2026)Ko, Lee, Chuang, Chatterjee, Shenoy, Lu, Moon, Dong,
  Bhardwaj, and Kim]{ko2026svmemagent}
Dohwan Ko, Ji~Soo Lee, Pierce Chuang, Debojeet Chatterjee, Ashish Shenoy,
  Yichao Lu, Seungwhan Moon, Xin~Luna Dong, Vikas Bhardwaj, and Hyunwoo~J Kim.
\newblock Svmemagent: A streaming video memory agent for query-agnostic online
  frame selection.
\newblock \emph{arXiv preprint arXiv:2609.18540}, 2026.

\bibitem[Kwon et~al.(2023)Kwon, Li, Zhuang, Sheng, Zheng, Yu, Gonzalez, Zhang,
  and Stoica]{kwon2023efficient}
Woosuk Kwon, Zhuohan Li, Siyuan Zhuang, Ying Sheng, Lianmin Zheng, Cody~Hao Yu,
  Joseph Gonzalez, Hao Zhang, and Ion Stoica.
\newblock Efficient memory management for large language model serving with
  pagedattention.
\newblock In \emph{Proceedings of the 29th symposium on operating systems
  principles}, pages 611--626, 2023.

\bibitem[Li et~al.(2024)Li, Zhang, Guo, Zhang, Li, Zhang, Zhang, Li, Liu, and
  Li]{li2024llavaone}
Bo~Li, Yuanhan Zhang, Dong Guo, Renrui Zhang, Feng Li, Hao Zhang, Kaichen
  Zhang, Yanwei Li, Ziwei Liu, and Chunyuan Li.
\newblock Llava-onevision: Easy visual task transfer.
\newblock \emph{arXiv preprint arXiv:2408.03326}, 2024.

\bibitem[Li et~al.(2025)Li, Tan, Shi, and Shao]{li2025videoscan}
Ruanjun Li, Yuedong Tan, Yuanming Shi, and Jiawei Shao.
\newblock Videoscan: Enabling efficient streaming video understanding via
  frame-level semantic carriers.
\newblock \emph{arXiv preprint arXiv:2503.09387}, 2025.

\bibitem[Liang et~al.(2026)Liang, Li, Chen, Zhang, Lu, and Li]{liang2026oasis}
Zhijia Liang, Jiaming Li, Weikai Chen, Yanhao Zhang, Haonan Lu, and Guanbin Li.
\newblock Oasis: On-demand hierarchical event memory for streaming video
  reasoning.
\newblock In \emph{Proceedings of the IEEE/CVF Conference on Computer Vision
  and Pattern Recognition (CVPR)}, 2026.

\bibitem[Lin et~al.(2024)Lin, Yin, Ping, Molchanov, Shoeybi, and
  Han]{lin2024vila}
Ji~Lin, Hongxu Yin, Wei Ping, Pavlo Molchanov, Mohammad Shoeybi, and Song Han.
\newblock Vila: On pre-training for visual language models.
\newblock In \emph{2024 IEEE/CVF Conference on Computer Vision and Pattern
  Recognition (CVPR)}, pages 26679--26689. IEEE, 2024.

\bibitem[Lin et~al.(2026{\natexlab{a}})Lin, Fang, Chen, Cheng, Wan, Luo, Wang,
  Li, Liu, and Sun]{lin2026streamingbench}
Junming Lin, Zheng Fang, Chi Chen, Haoxuan Cheng, Zihao Wan, Fuwen Luo, Ziyue
  Wang, Peng Li, Yang Liu, and Maosong Sun.
\newblock Streamingbench: Assessing the gap for mllms to achieve streaming
  video understanding.
\newblock In \emph{ICASSP 2026-2026 IEEE International Conference on Acoustics,
  Speech and Signal Processing (ICASSP)}, pages 12147--12151. IEEE,
  2026{\natexlab{a}}.

\bibitem[Lin et~al.(2026{\natexlab{b}})Lin, Tong, Wu, Zhang, Liu, Jin, and
  Shen]{lin2026speak}
Junyan Lin, Junlong Tong, Hao Wu, Jialiang Zhang, Jinming Liu, Xin Jin, and
  Xiaoyu Shen.
\newblock Speak while watching: Unleashing true real-time video understanding
  capability of multimodal large language models.
\newblock \emph{arXiv preprint arXiv:2601.06843}, 2026{\natexlab{b}}.

\bibitem[Liu et~al.(2026)Liu, Guo, Li, Zhen, He, Ji, Ren, Zhang, Lu, and
  Liu]{liu2026thinking}
Zikang Liu, Longteng Guo, Handong Li, Ru~Zhen, Xingjian He, Ruyi Ji, Xiaoming
  Ren, Yanhao Zhang, Haonan Lu, and Jing Liu.
\newblock Thinking in streaming video.
\newblock \emph{arXiv preprint arXiv:2603.12938}, 2026.

\bibitem[Lu et~al.(2026)Lu, Zhang, Tao, Qu, Li, Wan, and Wang]{lu2026vista}
Haocheng Lu, Nan Zhang, Wei Tao, Xiaoyang Qu, Guokuan Li, Jiguang Wan, and
  Jianzong Wang.
\newblock Vista: Scene-aware optimization for streaming video question
  answering under post-hoc queries.
\newblock In \emph{Proceedings of the AAAI Conference on Artificial
  Intelligence}, volume~40, pages 7539--7547, 2026.

\bibitem[Ning et~al.(2025)Ning, Liu, Jin, Li, Ding, Guo, and
  Zhao]{ning2025livevlm}
Zhenyu Ning, Guangda Liu, Qihao Jin, Chengwei Li, Wenchao Ding, Minyi Guo, and
  Jieru Zhao.
\newblock Livevlm: Efficient online video understanding via streaming-oriented
  kv cache and retrieval.
\newblock \emph{arXiv preprint arXiv:2505.15269}, 2025.

\bibitem[Niu et~al.(2025)Niu, Li, Miao, Ge, Zhou, He, Dong, Duan, Ding, Qian,
  et~al.]{niu2025ovo}
Junbo Niu, Yifei Li, Ziyang Miao, Chunjiang Ge, Yuanhang Zhou, Qihao He, Xiaoyi
  Dong, Haodong Duan, Shuangrui Ding, Rui Qian, et~al.
\newblock Ovo-bench: How far is your video-llms from real-world online video
  understanding?
\newblock In \emph{2025 IEEE/CVF Conference on Computer Vision and Pattern
  Recognition (CVPR)}, pages 18902--18913. IEEE, 2025.

\bibitem[OpenAI(2024)]{openai2024gpt4o}
OpenAI.
\newblock Gpt-4o system card, 2024.
\newblock URL \url{https://openai.com/index/hello-gpt-4o/}.

\bibitem[Qian et~al.(2025)Qian, Ding, Dong, Zhang, Zang, Cao, Lin, and
  Wang]{qian2025dispider}
Rui Qian, Shuangrui Ding, Xiaoyi Dong, Pan Zhang, Yuhang Zang, Yuhang Cao,
  Dahua Lin, and Jiaqi Wang.
\newblock Dispider: Enabling video llms with active real-time interaction via
  disentangled perception, decision, and reaction.
\newblock In \emph{2025 IEEE/CVF Conference on Computer Vision and Pattern
  Recognition (CVPR)}, pages 24045--24055. IEEE, 2025.

\bibitem[Qu et~al.(2026)Qu, Yao, Xing, Hu, Ding, Zhang, Zhang, Yuan, Shu, and
  Yan]{qu2026beyond}
Hongyu Qu, Guangming Yao, Ling Xing, Xiaobin Hu, Rongxing Ding, Guibin Zhang,
  Fan Zhang, Yi~Yuan, Xiangbo Shu, and Shuicheng Yan.
\newblock Beyond retrieval: Progressive latent memory evolution for streaming
  video understanding.
\newblock \emph{arXiv preprint arXiv:2609.04131}, 2026.

\bibitem[{Qwen Team}(2026)]{qwen3.6-27b}
{Qwen Team}.
\newblock {Qwen3.6-27B}: Flagship-level coding in a {27B} dense model, April
  2026.
\newblock URL \url{https://qwen.ai/blog?id=qwen3.6-27b}.

\bibitem[Radford et~al.(2023)Radford, Kim, Xu, Brockman, McLeavey, and
  Sutskever]{radford2023robust}
Alec Radford, Jong~Wook Kim, Tao Xu, Greg Brockman, Christine McLeavey, and
  Ilya Sutskever.
\newblock Robust speech recognition via large-scale weak supervision.
\newblock In \emph{International conference on machine learning}, pages
  28492--28518. PMLR, 2023.

\bibitem[Ravi et~al.(2025)Ravi, Gabeur, Hu, Hu, Ryali, Ma, Khedr, R{\"a}dle,
  Rolland, Gustafson, et~al.]{ravi2025sam}
Nikhila Ravi, Valentin Gabeur, Yuan-Ting Hu, Ronghang Hu, Chaitanya Ryali,
  Tengyu Ma, Haitham Khedr, Roman R{\"a}dle, Chloe Rolland, Laura Gustafson,
  et~al.
\newblock Sam 2: Segment anything in images and videos.
\newblock In \emph{International Conference on Learning Representations},
  volume 2025, pages 28085--28128, 2025.

\bibitem[Shen et~al.(2024)Shen, Xiong, Zhao, Wu, Chen, Zhu, Liu, Xiao,
  Varadarajan, Bordes, Liu, Xu, J.~Kim, Soran, Krishnamoorthi, Elhoseiny, and
  Chandra]{shen2024longvu}
Xiaoqian Shen, Yunyang Xiong, Changsheng Zhao, Lemeng Wu, Jun Chen, Chenchen
  Zhu, Zechun Liu, Fanyi Xiao, Balakrishnan Varadarajan, Florian Bordes, Zhuang
  Liu, Hu~Xu, Hyunwoo J.~Kim, Bilge Soran, Raghuraman Krishnamoorthi, Mohamed
  Elhoseiny, and Vikas Chandra.
\newblock Longvu: Spatiotemporal adaptive compression for long video-language
  understanding.
\newblock \emph{arXiv:2410.17434}, 2024.

\bibitem[Shen et~al.(2026)Shen, Tian, Yang, and Liu]{shen2026simple}
Yujiao Shen, Shulin Tian, Jingkang Yang, and Ziwei Liu.
\newblock A simple baseline for streaming video understanding.
\newblock \emph{arXiv preprint arXiv:2604.02317}, 2026.

\bibitem[Sitong et~al.(2026)Sitong, Yan, Kang, Zheng, Ruan, Lu, Zhang, Sato,
  and Huang]{sitong2026vinci2}
Gong Sitong, Tianyu Yan, Caixin Kang, Bo~Zheng, Xiang Ruan, Huchuan Lu, Kaipeng
  Zhang, Yoichi Sato, and Yifei Huang.
\newblock Vinci2: Providing proactive assistance in continuous egocentric
  videos.
\newblock \emph{arXiv preprint arXiv:2607.11523}, 2026.

\bibitem[Sun et~al.(2024)Sun, Yang, Zhang, Cheng, and Hou]{sun2023corrmatch}
Boyuan Sun, Yuqi Yang, Le~Zhang, Ming-Ming Cheng, and Qibin Hou.
\newblock Corrmatch: Label propagation via correlation matching for
  semi-supervised semantic segmentation.
\newblock \emph{IEEE Computer Vision and Pattern Recognition (CVPR)}, 2024.

\bibitem[Sun et~al.(2025{\natexlab{a}})Sun, Jin, Yin, and Hou]{sun2025depth}
Boyuan Sun, Modi Jin, Bowen Yin, and Qibin Hou.
\newblock Depth anything at any condition.
\newblock \emph{arXiv preprint arXiv:2507.01634}, 2025{\natexlab{a}}.

\bibitem[Sun et~al.(2025{\natexlab{b}})Sun, Zhao, Wei, and Hou]{sun2025llava}
Boyuan Sun, Jiaxing Zhao, Xihan Wei, and Qibin Hou.
\newblock Llava-scissor: Token compression with semantic connected components
  for video llms.
\newblock \emph{arXiv preprint arXiv:2506.21862}, 2025{\natexlab{b}}.

\bibitem[Sun et~al.(2026)Sun, Yin, Li, Wei, and Hou]{sun2026see}
Boyuan Sun, Bo-Wen Yin, Yuan-Ming Li, Xihan Wei, and Qibin Hou.
\newblock See what i mean: Aligning vision and language representations for
  video fine-grained object understanding.
\newblock In \emph{Proceedings of the IEEE/CVF Conference on Computer Vision
  and Pattern Recognition}, pages 36814--36827, 2026.

\bibitem[Tao et~al.(2025)Tao, Liao, Chen, Yin, Zhang, Liu, and
  Wang]{tao2025infinitevl}
Hongyuan Tao, Bencheng Liao, Shaoyu Chen, Haoran Yin, Qian Zhang, Wenyu Liu,
  and Xinggang Wang.
\newblock Infinitevl: Synergizing linear and sparse attention for
  highly-efficient, unlimited-input vision-language models.
\newblock \emph{arXiv preprint}, 2025.

\bibitem[Team et~al.(2024)Team, Georgiev, Lei, Burnell, Bai, Gulati, Tanzer,
  Vincent, Pan, Wang, et~al.]{team2024gemini}
Gemini Team, Petko Georgiev, Ving~Ian Lei, Ryan Burnell, Libin Bai, Anmol
  Gulati, Garrett Tanzer, Damien Vincent, Zhufeng Pan, Shibo Wang, et~al.
\newblock Gemini 1.5: Unlocking multimodal understanding across millions of
  tokens of context.
\newblock \emph{arXiv preprint arXiv:2403.05530}, 2024.

\bibitem[Team et~al.(2025)Team, Kamath, Ferret, Pathak, Vieillard, Merhej,
  Perrin, Matejovicova, Ram{\'e}, Rivi{\`e}re, et~al.]{team2025gemma}
Gemma Team, Aishwarya Kamath, Johan Ferret, Shreya Pathak, Nino Vieillard,
  Ramona Merhej, Sarah Perrin, Tatiana Matejovicova, Alexandre Ram{\'e},
  Morgane Rivi{\`e}re, et~al.
\newblock Gemma 3 technical report.
\newblock \emph{arXiv preprint arXiv:2503.19786}, 2025.

\bibitem[team(2024)]{Qwen2-VL}
Qwen team.
\newblock Qwen2-vl.
\newblock 2024.

\bibitem[Wang et~al.(2026{\natexlab{a}})Wang, Feng, Lai, Xu, Li, Ge, Dehghan,
  Cao, and Huang]{wang2026streambridge}
Haibo Wang, Bo~Feng, Zhengfeng Lai, Mingze Xu, Shiyu Li, Weifeng Ge, Afshin
  Dehghan, Meng Cao, and Ping Huang.
\newblock Streambridge: Turning your offline video large language model into a
  proactive streaming assistant.
\newblock \emph{Advances in Neural Information Processing Systems},
  38:\penalty0 132332--132359, 2026{\natexlab{a}}.

\bibitem[Wang et~al.(2026{\natexlab{b}})Wang, Sun, Zhu, Li, Xu, Wen, Hu, Li,
  and Zhang]{wang2026streammeco}
Junxi Wang, Te~Sun, Jiayi Zhu, Junxian Li, Haowen Xu, Zichen Wen, Xuming Hu,
  Zhiyu Li, and Linfeng Zhang.
\newblock Streammeco: Long-term agent memory compression for efficient
  streaming video understanding.
\newblock In \emph{Findings of the Association for Computational Linguistics:
  ACL 2026}, pages 13234--13251, 2026{\natexlab{b}}.

\bibitem[Wang et~al.(2026{\natexlab{c}})Wang, Jin, Hao, Chen, Liu, Ao, and
  Zhao]{wang2026think}
Lu~Wang, Zhuoran Jin, Yupu Hao, Yubo Chen, Kang Liu, Yulong Ao, and Jun Zhao.
\newblock Think while watching: Online streaming segment-level memory for
  multi-turn video reasoning in multimodal large language models.
\newblock \emph{arXiv preprint arXiv:2603.11896}, 2026{\natexlab{c}}.

\bibitem[Wang et~al.(2023)Wang, Zhu, Zhang, Huang, Ye, Xu, Chen, and
  Wang]{wang2023we}
Xiaofeng Wang, Zheng Zhu, Yunpeng Zhang, Guan Huang, Yun Ye, Wenbo Xu, Ziwei
  Chen, and Xingang Wang.
\newblock Are we ready for vision-centric driving streaming perception? the
  asap benchmark.
\newblock In \emph{2023 IEEE/CVF Conference on Computer Vision and Pattern
  Recognition (CVPR)}, pages 9600--9610. IEEE, 2023.

\bibitem[Wang et~al.(2026{\natexlab{d}})Wang, Liu, Gui, Lin, Yang, Liao, Chen,
  and Zhang]{wang2026accelerating}
Yiyu Wang, Xuyang Liu, Xiyan Gui, Xinying Lin, Boxue Yang, Chenfei Liao, Tailai
  Chen, and Linfeng Zhang.
\newblock Accelerating streaming video large language models via hierarchical
  token compression.
\newblock In \emph{Proceedings of the IEEE/CVF Conference on Computer Vision
  and Pattern Recognition}, pages 18523--18533, 2026{\natexlab{d}}.

\bibitem[Wang et~al.(2025)Wang, Wang, Chen, Wu, Zhao, and
  Zheng]{wang2025omnimmi}
Yuxuan Wang, Yueqian Wang, Bo~Chen, Tong Wu, Dongyan Zhao, and Zilong Zheng.
\newblock Omnimmi: A comprehensive multi-modal interaction benchmark in
  streaming video contexts.
\newblock In \emph{2025 IEEE/CVF Conference on Computer Vision and Pattern
  Recognition (CVPR)}, pages 18925--18935. IEEE, 2025.

\bibitem[Wei et~al.(2026)Wei, Huang, Chen, Hou, and Qi]{wei2026see}
Yuxi Wei, Wei Huang, Qirui Chen, Lu~Hou, and Xiaojuan Qi.
\newblock See, remember, explore: A benchmark and baselines for streaming
  spatial reasoning.
\newblock \emph{arXiv preprint arXiv:2603.23864}, 2026.

\bibitem[Wen et~al.(2026)Wen, Wang, Zhang, Huang, and Wu]{wen2026eventmemagent}
Siwei Wen, Zhangcheng Wang, Xingjian Zhang, Lei Huang, and Wenjun Wu.
\newblock Eventmemagent: Hierarchical event-centric memory for online video
  understanding with adaptive tool use.
\newblock In \emph{European Conference on Computer Vision (ECCV)}, 2026.

\bibitem[Xia et~al.(2026)Xia, Chen, Zhang, Sun, and Zhou]{xia2026streaming}
Jiaer Xia, Peixian Chen, Mengdan Zhang, Xing Sun, and Kaiyang Zhou.
\newblock Streaming video instruction tuning.
\newblock In \emph{Proceedings of the IEEE/CVF Conference on Computer Vision
  and Pattern Recognition}, pages 31219--31229, 2026.

\bibitem[Xiong et~al.(2025)Xiong, Yang, Yu, Zhuge, Zhang, Zhu, and
  Lu]{xiong2025streaming}
Haomiao Xiong, Zongxin Yang, Jiazuo Yu, Yunzhi Zhuge, Lu~Zhang, Jiawen Zhu, and
  Huchuan Lu.
\newblock Streaming video understanding and multi-round interaction with
  memory-enhanced knowledge.
\newblock In \emph{International Conference on Learning Representations},
  volume 2025, pages 69332--69351, 2025.

\bibitem[Yang et~al.(2025{\natexlab{a}})Yang, Tang, Zhao, Zhuang, Lu, An, Hu,
  Zhang, Swikir, He, et~al.]{yang2025streamagent}
Haolin Yang, Feilong Tang, Lingxiao Zhao, Xinlin Zhuang, Yifan Lu, Xiang An,
  Ming Hu, Xiaofeng Zhang, Abdalla Swikir, Junjun He, et~al.
\newblock Streamagent: Towards anticipatory agents for streaming video
  understanding.
\newblock \emph{arXiv preprint arXiv:2508.01875}, 2025{\natexlab{a}}.

\bibitem[Yang et~al.(2025{\natexlab{b}})Yang, Yao, Chen, Fu, Bai, Zhao, Sun,
  Yin, Wei, and Zhou]{yang2025humanomniv2}
Qize Yang, Shimin Yao, Weixuan Chen, Shenghao Fu, Detao Bai, Jiaxing Zhao,
  Boyuan Sun, Bowen Yin, Xihan Wei, and Jingren Zhou.
\newblock Humanomniv2: From understanding to omni-modal reasoning with context.
\newblock \emph{arXiv preprint arXiv:2506.21277}, 2025{\natexlab{b}}.

\bibitem[Yang et~al.(2026)Yang, Zhang, Jia, Guo, Shen, Zhang, Zhang, Wang, Li,
  Zhang, et~al.]{yang2026mage}
Senqiao Yang, Kaichen Zhang, Zhaoyang Jia, Jinghao Guo, Yifei Shen, Xinjie
  Zhang, Xiaoyi Zhang, Haoqing Wang, Xiao Li, Peng Zhang, et~al.
\newblock Mage-vl: An efficient codec-native streaming multimodal foundation
  model.
\newblock \emph{arXiv preprint arXiv:2607.24904}, 2026.

\bibitem[Yang et~al.(2025{\natexlab{c}})Yang, Zhao, Shukla, Singh, Mishra,
  Zhang, and Ren]{yang2025streammem}
Yanlai Yang, Zhuokai Zhao, Satya~Narayan Shukla, Aashu Singh, Shlok~Kumar
  Mishra, Lizhu Zhang, and Mengye Ren.
\newblock Streammem: Query-agnostic kv cache memory for streaming video
  understanding.
\newblock \emph{arXiv preprint arXiv:2508.15717}, 2025{\natexlab{c}}.

\bibitem[Yang et~al.(2025{\natexlab{d}})Yang, Hu, Du, Xue, Qian, Wu, Yang,
  Dong, and Xu]{yang2025svbench}
Zhenyu Yang, Yuhang Hu, Zemin Du, Dizhan Xue, Shengsheng Qian, Jiahong Wu, Fan
  Yang, Weiming Dong, and Changsheng Xu.
\newblock Svbench: A benchmark with temporal multi-turn dialogues for streaming
  video understanding.
\newblock In \emph{International Conference on Learning Representations},
  volume 2025, pages 42522--42556, 2025{\natexlab{d}}.

\bibitem[Yao et~al.(2026)Yao, Zhou, Yang, Qin, Hou, Liang, Wang, Cao, Ye, Xie,
  et~al.]{yao2026joyai}
Dingyu Yao, Junhao Zhou, Chenxu Yang, Chuanyu Qin, Haowen Hou, Zheming Liang,
  Congcong Wang, Yuhang Cao, Shenglong Ye, Shuai Xie, et~al.
\newblock Joyai-vl-interaction: Real-time vision-language interaction
  intelligence.
\newblock \emph{arXiv preprint arXiv:2606.14777}, 2026.

\bibitem[Yao et~al.(2025)Yao, Li, Wei, Li, Ren, Liu, Ouyang, Wang, Li, Li,
  et~al.]{yao2025timechat}
Linli Yao, Yicheng Li, Yuancheng Wei, Lei Li, Shuhuai Ren, Yuanxin Liu, Kun
  Ouyang, Lean Wang, Shicheng Li, Sida Li, et~al.
\newblock Timechat-online: 80\% visual tokens are naturally redundant in
  streaming videos.
\newblock In \emph{Proceedings of the 33rd ACM International Conference on
  Multimedia}, pages 10807--10816, 2025.

\bibitem[Zeng et~al.(2026)Zeng, Qiu, Zhang, Li, Wang, Li, Yan, Tian, Tian,
  Zhao, et~al.]{zeng2026streamforest}
Xiangyu Zeng, Kefan Qiu, Qingyu Zhang, Xinhao Li, Jing Wang, Jiaxin Li, Ziang
  Yan, Kun Tian, Meng Tian, Xinhai Zhao, et~al.
\newblock Streamforest: Efficient online video understanding with persistent
  event memory.
\newblock \emph{Advances in Neural Information Processing Systems},
  38:\penalty0 75804--75835, 2026.

\bibitem[Zhang et~al.(2026{\natexlab{a}})Zhang, Bi, He, Zhang, Lin, Xiao, Fu,
  Xie, Xie, Chen, et~al.]{zhang2026streamscout}
Ce~Zhang, Jing Bi, Jinxi He, Jianshu Zhang, Jingyang Lin, Yunzhong Xiao,
  Minghao Fu, Yaqi Xie, Zhentao Xie, Weicong Chen, et~al.
\newblock Streamscout: Learning when to look deeper for streaming video
  understanding.
\newblock \emph{arXiv preprint arXiv:2609.00291}, 2026{\natexlab{a}}.

\bibitem[Zhang et~al.(2025{\natexlab{a}})Zhang, Ding, Liu, Zhang, and
  Tresp]{zhang2025memory}
Gengyuan Zhang, Mingcong Ding, Tong Liu, Yao Zhang, and Volker Tresp.
\newblock Memory helps, but confabulation misleads: Understanding streaming
  events in videos with mllms.
\newblock \emph{arXiv preprint arXiv:2502.15457}, 2025{\natexlab{a}}.

\bibitem[Zhang et~al.(2024{\natexlab{a}})Zhang, Wang, Tang, Liu, Feng, Dai, and
  Jin]{zhang2024flashvstream}
Haoji Zhang, Yiqin Wang, Yansong Tang, Yong Liu, Jiashi Feng, Jifeng Dai, and
  Xiaojie Jin.
\newblock Flash-vstream: Memory-based real-time understanding for long video
  streams.
\newblock \emph{arXiv preprint arXiv:2406.08085}, 2024{\natexlab{a}}.

\bibitem[Zhang et~al.(2025{\natexlab{b}})Zhang, Wang, Tang, Liu, Feng, and
  Jin]{zhang2025flash}
Haoji Zhang, Yiqin Wang, Yansong Tang, Yong Liu, Jiashi Feng, and Xiaojie Jin.
\newblock Flash-vstream: Efficient real-time understanding for long video
  streams.
\newblock In \emph{2025 IEEE/CVF International Conference on Computer Vision
  (ICCV)}, pages 21059--21069. IEEE, 2025{\natexlab{b}}.

\bibitem[Zhang et~al.(2026{\natexlab{b}})Zhang, Yang, Fu, Ng, and
  Qiu]{zhang2026hermes}
Haowei Zhang, Shudong Yang, Jinlan Fu, See~Kiong Ng, and Xipeng Qiu.
\newblock Hermes: Kv cache as hierarchical memory for efficient streaming video
  understanding.
\newblock In \emph{Proceedings of the 64th Annual Meeting of the Association
  for Computational Linguistics (Volume 1: Long Papers)}, pages 8411--8430,
  2026{\natexlab{b}}.

\bibitem[Zhang et~al.(2026{\natexlab{c}})Zhang, Yang, Wang, Qian, and
  Xu]{zhang2026querystream}
Kairui Zhang, Zhenyu Yang, Bing Wang, Shengsheng Qian, and Changsheng Xu.
\newblock Querystream: Advancing streaming video understanding with query-aware
  pruning and proactive response.
\newblock In \emph{The Fourteenth International Conference on Learning
  Representations}, 2026{\natexlab{c}}.

\bibitem[Zhang et~al.(2024{\natexlab{b}})Zhang, Zhang, Li, Zeng, Yang, Zhang,
  Wang, Tan, Li, and Liu]{zhang2024long}
Peiyuan Zhang, Kaichen Zhang, Bo~Li, Guangtao Zeng, Jingkang Yang, Yuanhan
  Zhang, Ziyue Wang, Haoran Tan, Chunyuan Li, and Ziwei Liu.
\newblock Long context transfer from language to vision.
\newblock \emph{arXiv preprint arXiv:2406.16852}, 2024{\natexlab{b}}.

\bibitem[Zhang et~al.(2024{\natexlab{c}})Zhang, Wu, Li, Li, Ma, Liu, and
  Li]{zhang2024videoinstructiontuningsynthetic}
Yuanhan Zhang, Jinming Wu, Wei Li, Bo~Li, Zejun Ma, Ziwei Liu, and Chunyuan Li.
\newblock Video instruction tuning with synthetic data, 2024{\natexlab{c}}.
\newblock URL \url{https://arxiv.org/abs/2410.02713}.

\bibitem[Zhang et~al.(2026{\natexlab{d}})Zhang, Shi, and
  Yang]{zhang2026weavetime}
Yulin Zhang, Cheng Shi, and Sibei Yang.
\newblock Weavetime: Stream from earlier frames into emergent memory in
  videollms.
\newblock \emph{arXiv preprint arXiv:2602.22142}, 2026{\natexlab{d}}.

\bibitem[Zhao et~al.(2025{\natexlab{a}})Zhao, Sun, Chen, Wei, and
  Hou]{zhao2025llava}
Jiaxing Zhao, Boyuan Sun, Xiang Chen, Xihan Wei, and Qibin Hou.
\newblock Llava-octopus: Unlocking instruction-driven adaptive projector fusion
  for video understanding.
\newblock \emph{arXiv preprint arXiv:2501.05067}, 2025{\natexlab{a}}.

\bibitem[Zhao et~al.(2025{\natexlab{b}})Zhao, Yang, Peng, Bai, Yao, Sun, Chen,
  Fu, Wei, Bo, et~al.]{zhao2025humanomni}
Jiaxing Zhao, Qize Yang, Yixing Peng, Detao Bai, Shimin Yao, Boyuan Sun, Xiang
  Chen, Shenghao Fu, Xihan Wei, Liefeng Bo, et~al.
\newblock Humanomni: A large vision-speech language model for human-centric
  video understanding.
\newblock \emph{arXiv preprint arXiv:2501.15111}, 2025{\natexlab{b}}.

\bibitem[Zhao et~al.(2026{\natexlab{a}})Zhao, Sun, Chen, and
  Wei]{zhao2026facial}
Jiaxing Zhao, Boyuan Sun, Xiang Chen, and Xihan Wei.
\newblock Facial dynamics in video: Instruction tuning for improved facial
  expression perception and contextual awareness.
\newblock In \emph{Proceedings of the AAAI Conference on Artificial
  Intelligence}, volume~40, pages 13163--13171, 2026{\natexlab{a}}.

\bibitem[Zhao et~al.(2026{\natexlab{b}})Zhao, Wang, Sun, Zhou, Wu, Song, Shi,
  Liu, Zhu, Zhou, et~al.]{zhao2026realtime}
Ruixiang Zhao, Hualei Wang, Renhe Sun, Enzhi Zhou, Jincenzi Wu, Xujie Song,
  Kexin Shi, Zihang Liu, Pengcheng Zhu, Jiayi Zhou, et~al.
\newblock Realtime-venus: A full-duplex interaction system with asynchronous
  delegation.
\newblock \emph{arXiv preprint arXiv:2609.13814}, 2026{\natexlab{b}}.

\bibitem[Zhao et~al.(2026{\natexlab{c}})Zhao, Yang, Xin, Wang, Rao, LYU, and
  Li]{zhao2026omnipro}
Ruixiang Zhao, Jie Yang, Zijie Xin, Tianyi Wang, Fengyun Rao, Jing LYU, and
  Xirong Li.
\newblock Omnipro: A comprehensive benchmark for omni-proactive streaming video
  understanding.
\newblock \emph{arXiv preprint arXiv:2605.18577}, 2026{\natexlab{c}}.

\bibitem[Zheng et~al.(2026)Zheng, Ding, Yang, Jiang, Wu, Zhang, Wang, Cao, and
  Liu]{zheng2026garde}
Yikai Zheng, Xin Ding, Yifan Yang, Shiqi Jiang, Hao Wu, Qianxi Zhang, Weijun
  Wang, Ting Cao, and Yunxin Liu.
\newblock Em-garde: A propose-match framework for proactive streaming video
  understanding.
\newblock \emph{arXiv preprint arXiv:2603.19054}, 2026.

\bibitem[Zhu et~al.(2025)Zhu, Wang, Chen, Liu, Ye, Gu, Tian, Duan, Su, Shao,
  et~al.]{zhu2025internvl3}
Jinguo Zhu, Weiyun Wang, Zhe Chen, Zhaoyang Liu, Shenglong Ye, Lixin Gu, Hao
  Tian, Yuchen Duan, Weijie Su, Jie Shao, et~al.
\newblock Internvl3: Exploring advanced training and test-time recipes for
  open-source multimodal models.
\newblock \emph{arXiv preprint arXiv:2504.10479}, 2025.

\end{thebibliography}
\bibliographystyle{plainnat}

\clearpage
\appendix
\section*{\LARGE Appendix for DeltaStreamer: Towards Human-Like Agentic Streaming Video Understanding}

These supplementary materials document implementation rules and parameter settings, benchmark task definitions and evaluation details, memory and tool-use statistics, the prompt templates used by DeltaStreamer, and additional qualitative examples. We also discuss limitations and broader impacts in the discussion section.

\begingroup
\color{black}
\hypersetup{linkbordercolor=black,linkcolor=black,pdfborder={0 0 0},linktoc=all}
\fontsize{11}{22.5}\selectfont
\etoctoccontentsline{part}{Appendix}
\etocsettocdepth{subsubsection}
\localtableofcontents
\endgroup

\newpage
\section{Methodology Details}
\label{app:implementation}

\subsection{Observation Scoring Details}
\label{app:observation}

For the novelty score in Eq.~\ref{eq:novelty}, the history window contains up to 12 recent observations. Field tokens are extracted using regular-expression matching, converted to lowercase, and deduplicated. The field weights are 0.35 for delta, 0.25 for actions, 0.15 for actors, 0.10 for objects, 0.10 for OCR, and 0.05 for scene; description receives the default weight of 0.05. Delta receives the largest individual weight to prioritize explicit visual changes when assessing new information. With no observation history, a nonempty observation receives novelty 1. A field without a matching historical token set has maximum similarity 0, and an observation with no nonempty token set receives novelty 0.

Salience converts novelty into a weight for later retrieval, increasing the priority of observations that add information beyond recurring scene descriptions. For novelty $s\geq0.3$, salience is $1+0.5s$ and the observation is marked as change-bearing; otherwise, salience is 0.5. A separate repeated-delta guard demotes repeated templates to salience 0.5 and clears the change flag. This decision uses the full structured observation and repetition checks, including when the delta reports a verified-static frame. The associated change flag also determines whether a group of observations satisfies the change requirement for episodic aggregation.

\subsection{Memory Construction and Rendering}
\label{app:memory}

\paragraph{Stored records and cross-layer links.}
Working events retain their timestamp, observation text, delta, salience, change flag, referenced frame, and optional parent episode identifier. Episodes store their time span, summary, and supporting working-event identifiers. Long-term pins retain their source episode identifier and are created from model-emitted recap blocks or automatically promoted episode boundaries. Separate reverse pointers support navigation in both directions without introducing cycles into the hierarchy's directed abstraction edges. An episode need not have a long-term pin to support working--episodic retrieval.

\subsubsection{Episode Construction}
Episodes can be created as new observations arrive or when memory is prepared for a model call. Online boundary detection identifies when to close the current episode, while aggregation and coverage rules summarize observations that have not yet been assigned to one. The model can also request an explicit episode through a Recap.

\textbf{Online boundaries.}
A run is the sequence of working observations accumulated since the previous online boundary. Bayesian Online Changepoint Detection (BOCPD)~\citep{adams2007bayesian} estimates whether an incoming observation continues this sequence or begins a new one. A boundary is proposed when the most probable hypothesis has index zero, indicating a new sequence, and the accumulated run contains at least three observations. Before accepting it, a content-continuity check compares the incoming observation with the first observation of the run. If their scene descriptions match and their estimated actor counts agree, with both fields available, the boundary is suppressed to avoid splitting a continuous scene. Otherwise, the accumulated observations are summarized and linked to an episode, the detector is reset, and the incoming observation starts the next run.

\textbf{Aggregation and coverage.}
Some working events are not assigned to an episode by online boundary detection. When rendering memory, the system groups these uncovered events chronologically, starting a new group when consecutive events are more than 10 seconds apart or including the next event would extend the group beyond 45 seconds. Groups with at least five events and one change-bearing observation are retained. Adjacent retained groups are then merged when their mean field-level Jaccard similarity is at least 0.85 and their temporal gap is at most 30 seconds. This consolidates observations that describe similar content; the 45-second limit applies before merging.

The change requirement can leave long, mostly static intervals without an episodic summary, even though their observations remain in working memory. A coverage fallback therefore checks the interval after the latest episode and before the recent observations still shown individually in visible memory. If this interval spans at least 30 seconds, uncovered events are split into initial groups of at most 45 seconds, with at least two events per group, and the same similarity-based merging is applied. These groups become anchor episodes with lower base salience, preserving temporal coverage without requiring a detected change. A well-formed model-emitted $\langle\text{Recap}\rangle$ also creates an explicit episode and a corresponding long-term pin.

\textbf{BOCPD update.}
Given the structured observation $O_t$ at index $t$ and $\mathbf{x}_t=\phi(O_t)$ its embedding. The function $\phi$ hashes lowercase tokens into a 512-dimensional count vector and applies L2 normalization. The implementation updates the run-hypothesis probabilities as
\begin{equation}
\widetilde{p}_t(0)=H\,p_{\mathrm{prior}}(\mathbf{x}_t),\qquad
\widetilde{p}_t(r+1)=(1-H)\,p_{t-1}(r)\,
p(\mathbf{x}_t\mid r),
\label{eq:bocpd-update}
\end{equation}
Here, $t$ is the current observation index and $\mathbf{x}_t$ is its embedding vector. $H$ is the constant probability of starting a new run, and $r$ is the nonnegative index of a previous run hypothesis. In the implementation's indexing convention, hypothesis $r$ contains $r+1$ observations before the update, and new-run index zero contains the current observation alone. $p_{t-1}(r)$ is the normalized probability of the previous hypothesis; $p(\mathbf{x}_t\mid r)$ is its predictive density for the new embedding; and $p_{\mathrm{prior}}(\mathbf{x}_t)$ is the new-run predictive density. $\widetilde p_t(0)$ and $\widetilde p_t(r+1)$ are unnormalized new-run and continuation probabilities, respectively. Normalizing over retained hypotheses gives the current probability distribution $p_t$. 
The predictive distribution for a continuing run is
\begin{equation}
p(\mathbf{x}_t\mid r)=
\mathcal{N}\!\left(
\mathbf{x}_t;
\frac{\mathbf{s}_r}{n_r+\kappa_0},
\left(1+\frac{1}{n_r+\kappa_0}\right)\mathbf{I}_d
\right),
\qquad \kappa_0=0.1,
\label{eq:bocpd-predictive}
\end{equation}
where $p(\mathbf{x}_t\mid r)$ is the density of the current observation embedding $\mathbf{x}_t$ under previous run hypothesis $r$, and $t$ is the observation index. $\mathcal{N}$ denotes a multivariate Gaussian density whose arguments are the evaluation vector, mean, and covariance, in that order. $n_r$ is the number of observations in hypothesis $r$, $\mathbf{s}_r$ is the sum of their embedding vectors, $\kappa_0$ is the prior mean pseudo-count, and $\mathbf{I}_d$ is the identity matrix in embedding dimension $d$.

For a new run, the prior predictive density is $p_{\mathrm{prior}}(\mathbf{x}_t)=\mathcal{N}(\mathbf{x}_t;\mathbf{0}_d,(1+1/\kappa_0)\mathbf{I}_d)$. Here, $\mathbf{x}_t$ is the current embedding, $\mathcal{N}$ is the Gaussian density, $\mathbf{0}_d$ and $\mathbf{I}_d$ are the zero vector and identity matrix in embedding dimension $d$, and $\kappa_0=0.1$ is the same prior pseudo-count. Thus, the implementation incrementally updates the run mean but uses a fixed unit observation variance with posterior-mean uncertainty, rather than estimating a separate empirical variance for every run. At most 500 run hypotheses are retained.

\subsubsection{Entity Index and Memory Rendering}

The entity index records each entity's working-event identifiers, first and most recent timestamps, and observed surface names. Entities are extracted from observation text by rule-based matching. Cross-frame references are linked conservatively through name normalization, including matching a generic head noun to a more specific compound name, such as ``man'' and ``man in white.'' A compact digest exposes frequently mentioned entities and their temporal spans.

\textbf{Evidence paths.}
Among query-matching records, $\operatorname{Backtrack}$ in Eq.~\ref{eq:memory} selects at most two long-term pins, three episodes, and three working events as starting nodes, excluding question pins and planner notes. Source expansion includes at most three working events per episode. Paths receive a ranking bonus for covering multiple memory layers, are deduplicated by their starting node, and are returned in descending score order up to the requested limit. The layer quotas allow at most eight paths.

\textbf{Visible memory.}
$\operatorname{Recall}$ in Eq.~\ref{eq:memory} ranks keyword-matching working records using salience, a change bonus, and recency, with an additional token-hash similarity score when cached embeddings are available. The renderer takes the highest-ranked candidates and retains those in the middle interval, excluding entries already covered by the visible head or recent tail. The selected observations are displayed chronologically. Because this filtering follows retrieval, the recalled set can contain fewer records than the requested limit.


\subsection{Planner Schema and Evidence Guidance}
\label{app:planner}

The implementation uses the following fields for the plan in Eq.~\ref{eq:planner}.

\begin{center}
\small
\renewcommand{\arraystretch}{1.15}
\begin{tabularx}{\textwidth}{@{}p{0.18\textwidth}XX@{}}
\toprule
\textbf{Field} & \textbf{Meaning} & \textbf{Use in investigation} \\
\midrule
\texttt{need} & Target and its temporal anchor. & Forms the retrieval query. \\
\texttt{pov} & Viewpoint of the relevant participants. & First/third-person references or actors. \\
\texttt{info\_type} & The evidence needed, such as an action order, count, or attribute. & Guides the choice of observations and tool use. \\
\texttt{evidence\_locus} & Where the decisive evidence lies in time. & Guides retrieval of past evidence, current inspection, or continued observation. \\
\texttt{answerability} & Whether the listed options include an inability-to-answer choice. & Guides target-existence checks and selection among the permitted answers. \\
\texttt{meta} & Whether the query concerns answerability or an event's start or completion. & Distinguishes evidence of an event's state from a description of the event itself. \\
\texttt{strategy} & The proposed sequence of memory access and tool use. & Provides an initial course of action that the agent can adapt to new evidence. \\
\bottomrule
\end{tabularx}
\end{center}

Query construction removes answer-option text, common instruction words, and duplicate tokens from the question and the \texttt{need} field. Rule-based matching interprets the free-form temporal scope, evidence type, and strategy to derive an evidence category and tool-use guidance. 
Appendix~\ref{app:prompt-planner} provides the planning instruction and the runtime template for persistent plan fields and the evidence contract.

\subsection{Evidence-Gathering Tools}
\label{app:tools}

We expose each tool through a shared description template specifying its name, purpose, and JSON argument schema. To request execution, the agent emits a \texttt{<tool\_call>} block with \texttt{name} and \texttt{arguments} fields, then waits for the result before continuing. The shared call template and instructions for interpreting results and selecting subsequent tools are reproduced in Appendix~\ref{app:prompt-evidence}.

The tools in Tab.~\ref{tab:tools} use SAM2~\citep{ravi2025sam} for object tracking and Whisper~\citep{radford2023robust} for speech transcription. Post-query model calls receive a single current frame, with additional images supplied by spatial tools. Tool rounds can continue at the same frame, whereas continued observation advances to new frames. Interpretations of tool results also update memory.

Tools return a \texttt{<tool\_result>} block containing their name, success status, and result or error. A successful zoom additionally supplies the crop as an image input to the following model call. Replay returns the selected replay time and schedules historical-frame inspection through the temporal controller. When the seek is applied, the historical frame becomes the visual input, and the previous stream position is retained for return.

\begin{table}[htbp]
\centering
\small
\caption{Tool suite of DeltaStreamer, grouped by function.}
\label{tab:tools}
\setlength{\tabcolsep}{6pt}
\begin{tabular}{@{}>{\raggedright\arraybackslash}p{50pt}>{\raggedright\arraybackslash}p{90pt}>{\raggedright\arraybackslash}p{\dimexpr\linewidth-164pt\relax}@{}}
\toprule
\textbf{Group} & \textbf{Tool} & \textbf{Description} \\
\midrule
\multirow{4}{*}{\textbf{Memory}} & \texttt{memory\_search} & Cross-layer keyword retrieval over all memory DAG layers \\
& \texttt{memory\_read} & Retrieve specific events by query with time-range filtering \\
& \texttt{memory\_list} & List events filtered by kind and time range, chronologically \\
& \texttt{memory\_write} & Write a model-selected fact into memory \\
\midrule
\multirow{3}{*}[-2pt]{\textbf{Track}} & \texttt{track\_init} & Initialize a SAM2 object tracker on a box or prompt \\
& \texttt{track\_query} & Query the current position and confidence of a tracked object \\
& \texttt{track\_end} & Terminate a tracking session and obtain trajectory summary \\
\midrule
\textbf{Audio} & \texttt{asr\_segment} & Transcribe speech in a time window using Whisper ASR \\
\midrule
\textbf{Temporal} & \texttt{replay} & Schedule a past frame for visual inspection, retaining the current stream position for return. Recover visual details omitted from textual memory. \\
\midrule
\textbf{Spatial} & \texttt{zoom\_crop} & Crop and zoom into a sub-region of the current frame with IoU-based deduplication to prevent redundant zooming. \\
\midrule
\textbf{Efficiency} & \texttt{set\_sample\_rate} & Adjust the observation frame rate (0.25--2.0 fps) during static periods to save computation. \\
\bottomrule
\end{tabular}
\end{table}

\subsection{Look-Back Verification and Replay Details}
\label{app:verification}

Replaying historical frames incurs substantial overhead through additional frame access and visual processing, which can delay responses. We therefore use the checks and priority rules below to assess whether unresolved conflicts justify this cost before triggering automatic replay.

\textbf{Claim extraction and initial checks.}
When invoked, the verifier requests up to 16 evidence paths, subject to the layer quotas described in Appendix~\ref{app:memory}. Claim extraction uses fixed text patterns for explicit state tags, location phrases, a limited set of attributes, and numerical counts. Subject and property strings are lowercased, and claims about the same subject and property are grouped. The following preliminary rules are applied in order. The first matching rule skips subsequent cross-layer and temporal checks for that group and triggers no automatic replay. These outcomes concern the need for replay and do not establish that the proposed answer is visually correct.
\begin{enumerate}
    \item A group with fewer than two claims is marked uncertain and does not trigger replay.
    \item If working and episodic records contain the same set of values but long-term records differ, the group passes the initial check.
    \item If every gap between consecutive claims in temporal order is at most 2 seconds on the stored 1-fps timeline, the group passes the temporal-proximity check.
    \item If all claims have the same value, the group passes the agreement check.
    \item If at least 85\% of claims share one value, the group is marked uncertain without replay.
\end{enumerate}

\textbf{Conflict severity and target selection.}
Groups unresolved by the preliminary checks are examined for disagreement across memory layers and state transitions that may need visual confirmation. The cross-layer check gives greater weight to records closer to the original observations. The temporal check can flag transitions that may reflect missing intermediate evidence, even when the cross-layer check requests no replay. Either check can nominate a replay candidate, whose severity reflects the stronger signal. For a cross-layer conflict, the verifier targets the earliest supporting working observation when available. A flagged temporal transition supplies its earlier endpoint if no target is selected. The scheduler then ranks these candidates to decide whether and where to look back.

\textbf{Replay priority.}
For a conflict $u$ with severity $c_u$ and replay index $\tau_u$, the implemented priority in Eq.~\ref{eq:replay-selection} is
\begin{equation}
\rho(u)=\left[p_{\mathrm{rep}}-\max(0,1-c_u)\right]
\max\!\left(d_{\min},1-\lambda\left|\operatorname{time}(\tau_u)-\operatorname{time}(t_q)\right|\right),
\label{eq:replay-priority}
\end{equation}
Here, $u$ is an unresolved conflict, $\rho(u)$ is its replay priority, $c_u$ is its conflict-severity score, and $\tau_u$ is its target sampled-frame index. $t_q$ is the question-arrival frame index; $\operatorname{time}$ maps frame indices to video timestamps in seconds, and the absolute-value term is their temporal separation. $p_{\mathrm{rep}}$ is the scheduler's assumed correctness after replay; $\max(0,1-c_u)$ is its heuristic estimate without replay. $d_{\min}$ is the minimum temporal weighting factor and $\lambda$ is its decay rate per second. Defaults are $p_{\mathrm{rep}}=0.95$, $d_{\min}=0.5$, and $\lambda=0.02\,\mathrm{s}^{-1}$. 
Candidates with $\rho(u)\leq\eta$ are discarded, where $\eta=0.15$ is the minimum priority required to justify replay cost; the remainder are sorted by descending priority, deduplicated by target, and capped at three.

\section{Benchmark Details}
\label{app:benchmarks}

\textbf{Evaluation protocol.}
Query times beyond the supplied video segment are mapped to the last available frame. Planning and finalization calls can reuse the last available frame without exposing additional video. Accepted responses and their trigger times are read from the recorded outputs, and task-specific parsers extract the required answer format. Question and response times refer to video timestamps rather than wall-clock inference time.

\subsection{OVO-Bench}

OVO-Bench~\citep{niu2025ovo} contains 3035 evaluation samples spanning 12 tasks in three groups. The \textit{Real-Time Visual Perception} group (837 samples) includes Optical Character Recognition (OCR), Action Recognition (ACR), Attribute Recognition (ATR), Spatial Understanding (STU), Future Prediction (FPD), and Object Recognition (OJR). The \textit{Backward Tracing} group (631 samples) includes Episodic Memory (EPM), Action Sequence Identification (ASI), and Hallucination Detection (HLD). The \textit{Forward Active Responding} group (1567 samples) includes Repetition Event Count (REC), Sequential Steps Recognition (SSR), and Clues Reveal Responding (CRR).

\subsection{StreamingBench}

StreamingBench~\citep{lin2026streamingbench} contains 4500 QA samples in three groups. The \textit{Real-Time Visual Understanding} group covers Object Perception (OP), Causal Reasoning (CR), Clips Summarization (CS), Attribute Perception (ATP), Event Understanding (EU), Text-Rich Understanding (TR), Prospective Reasoning (PR), Spatial Understanding (SU), Action Perception (ACP), and Counting (CT). The \textit{Omni-Source Understanding} group covers Emotion Recognition (ER), Scene Understanding (SCU), Source Discrimination (SD), and Multimodal Alignment (MA). The \textit{Contextual Understanding} group covers Anomaly Context Understanding (ACU), Misleading Context Understanding (MCU), Sequential Question Answering (SQA), and Proactive Output (PO).


\section{Memory and Tool Use}
\label{app:runtime_statistics}

\begin{figure}[H]
\centering
\includegraphics[width=\textwidth]{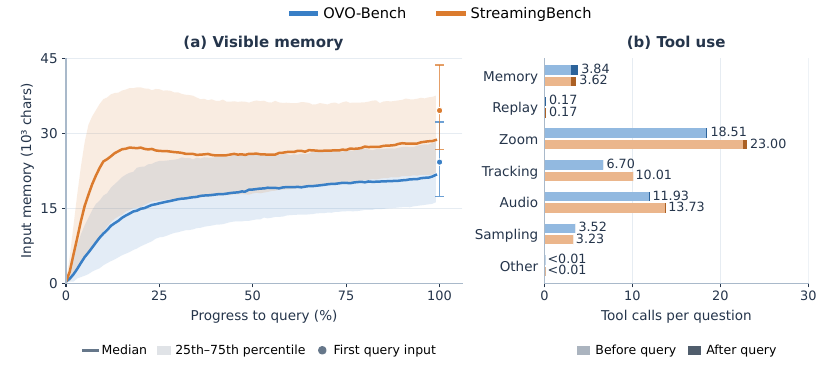}
\caption{Visible memory (a) and tool use (b) on OVO-Bench and StreamingBench.}
\label{fig:runtime_statistics}
\end{figure}

Fig.~\ref{fig:runtime_statistics} summarizes memory and tool-use patterns on OVO-Bench and StreamingBench. Blue denotes OVO-Bench, and orange denotes StreamingBench throughout the figure. The gray legend symbols describe plotting conventions shared by both benchmarks.

\textbf{Visible memory.}
We measure the length of the memory text supplied to the model and align each trajectory by its observation progress up to query arrival. In Fig.~\ref{fig:runtime_statistics}(a), solid lines show the median memory length across trajectories, and shaded bands span the 25th--75th percentiles. The circles at 100\% mark the median memory length in the first query-time model input, with vertical whiskers showing the corresponding 25th--75th percentile range. Memory grows quickly during the initial observation period, after which its growth slows substantially on OVO-Bench and largely stabilizes on StreamingBench. This pattern is consistent with selective rendering and episodic compression keeping the visible context under control as observations accumulate. The measurement concerns the rendered input memory, not the total storage of records and historical frames.

\textbf{Replay frequency.}
In Fig.~\ref{fig:runtime_statistics}(b), each tool has one bar per benchmark. Light and dark segments show mean dispatched calls per question before and after query arrival, respectively. The number at each bar end reports their sum. Low replay frequency is expected given its selective role in recovering evidence during question answering. Revisiting historical frames requires additional frame access and visual processing, so frequent replay can increase response overhead and interfere with real-time processing. The agent is instructed to answer from the current frame or retained memory when sufficient and to revisit a past moment when visual confirmation is needed. Replay is therefore concentrated in the answering phase, whereas observation tools are used throughout the stream. In the recorded trajectories, the last replay is typically followed by a short remaining sequence of agent steps before completion. This limited answering-stage window, together with tool budgets and termination rules, helps explain the low frequency of repeated replay and reflects the trade-off between additional inspection and timely responses.

\textbf{Observation, memory, and planning.}
Tab.~\ref{tab:structure_stats} summarizes the information recorded during observation and investigation. 
We count pre-query visual observations containing both \texttt{description} and \texttt{delta}. Text lengths are measured in Unicode characters; observation and delta lengths are averaged within each sample and then across samples with observations. Stored working records and episodes are counted at completion, with working records including observations, frame references, and tool-related entries. Plan length refers to the first generated plan.

\begin{table}[!t]
\centering
\small
\caption{Observation, memory, and planning statistics. All values are means.}
\label{tab:structure_stats}
\renewcommand{\arraystretch}{1.12}
\setlength{\tabcolsep}{0pt}
\setbox0=\hbox{\begin{tabular}{@{}lcc@{}}
Visual observations per sample & \textbf{OVO-Bench} & \textbf{StreamingBench}
\end{tabular}}
\setlength{\tabcolsep}{\dimexpr(\linewidth-\wd0)/4\relax}
\begin{tabular}{@{}lcc@{}}
\toprule
\textbf{Metric} & \textbf{OVO-Bench} & \textbf{StreamingBench} \\
\midrule
\multicolumn{3}{@{}c@{}}{\textbf{Observation before query}} \\
Visual observations per sample & 70.85 & 118.31 \\
Observation length (chars.) & 1,279.15 & 1,588.48 \\
Delta length (chars.) & 155.54 & 172.98 \\
\addlinespace[3pt]
\hline
\multicolumn{3}{@{}c@{}}{\textbf{Memory at completion}} \\
Stored working records & 131.91 & 261.71 \\
Episodes & 11.68 & 16.08 \\
\addlinespace[3pt]
\hline
\multicolumn{3}{@{}c@{}}{\textbf{Planning}} \\
Initial plan length (chars.) & 1,329.43 & 1,299.82 \\
\bottomrule
\end{tabular}
\end{table}

These statistics complement the visible-memory curves by describing what is recorded and retained. Delta occupies only part of each observation while making changes explicit. The memory retains detailed working records together with a smaller set of episodic summaries, and the initial plan states the evidence needs that guide subsequent investigation.

\section{Discussions}
\subsection{Limitations}
\label{app:limitations}

DeltaStreamer requires repeated model and tool calls, which may increase inference latency, although frame arrival and processing can proceed asynchronously in real-world streaming settings. Compact visible memory also leaves storage costs for records and historical frames.
Errors in observations or temporal anchors can affect retrieval, while agreement across memory layers does not ensure perfect correctness. Our evaluation uses one shared backbone on two benchmarks and leaves challenges in historical event localization, sequential-step recognition, and proactive response timing. The weaker PO results on StreamingBench highlight the additional challenge of precise response triggering, since retrospective evidence recovery cannot compensate for a missed response window. Further work should examine other backbones and the trade-offs among accuracy, latency, and storage.

\subsection{Broader Impacts}
\label{app:broader_impacts}

DeltaStreamer may support accessibility tools and help users review extended activities by retrieving relevant historical evidence. However, continuous observation, structured memory, and replay can expose personal routines and facilitate non-consensual monitoring. Errors or biases may propagate through stored descriptions and answers, misrepresenting people's actions. For deployment, we recommend informed consent from recorded individuals, limited retention, access controls, and deletion mechanisms covering both frames and derived memories. Human review of the underlying evidence should precede consequential decisions about individuals. These applications and the effectiveness of the recommended safeguards have not been evaluated in this work.


\clearpage

\begingroup
\fontencoding{T1}\fontfamily{ptm}\selectfont
\raggedbottom
\definecolor{dsPromptBack}{HTML}{F4F4F4}
\definecolor{dsPromptRule}{HTML}{777777}
\newtcolorbox{dsprompt}[2][]{
  breakable,
  colback=dsPromptBack,colframe=dsPromptRule,colbacktitle=dsPromptBack,coltitle=black,
  boxrule=0.5pt,arc=2mm,
  left=4mm,right=4mm,top=2mm,bottom=2mm,
  title={#2},title after break={#2 (continued)},titlerule=0pt,
  toptitle=2mm,bottomtitle=1mm,
  fonttitle=\fontencoding{T1}\fontfamily{ptm}\bfseries\fontsize{11}{13}\selectfont,
  fontupper=\fontencoding{T1}\fontfamily{ptm}\fontsize{10}{12.4}\selectfont,
  before upper={\setlength{\parindent}{0pt}\setlength{\parskip}{3.5pt}\raggedright},
  before skip=7pt,after skip=5pt,
  #1
}
\newcommand{\dsheading}[1]{\par\addvspace{7pt}\noindent\textbf{#1}\par\nobreak}
\newcommand{\dsfield}[2]{\par\addvspace{7pt}\noindent\textbf{#1}\hspace{0.35em}#2\par}
\newcommand{\dsplanfield}[2]{\par\addvspace{9pt}\noindent\textbf{#1}\par\nobreak #2\par}
\newcommand{\dsexample}[2]{{\leftskip=12pt\setlength{\parskip}{2pt}\noindent #1\hspace{0.3em}#2\par}}
\newcommand{\dsitem}[2]{{\hangindent=12pt\hangafter=1\noindent\makebox[12pt][l]{#1}#2\par}}
\newcommand{\dscode}[1]{{\leftskip=12pt\noindent #1\par}}

\section{Prompt Templates}
\label{app:prompts}

\subsection{Pre-Query Observation}
\label{app:prompt-observation}
The two-line instruction applies to the ordinary observation report followed by \texttt{</Silence>}. Tool calls occupy intermediate turns, after which the agent is prompted again on the same frame. A conditional \texttt{<Recap>} block may precede \texttt{</Silence>}, as described in Appendix~\ref{app:prompt-memory}.
The prompt requires a concrete change while forbidding invented changes, with exceptions for the initial frame and verified identical frames. 

\begin{dsprompt}[]{Observation prompt}
[Phase] OBSERVE-ONLY. The question has not been asked yet.\par
Reply with EXACTLY two lines:\par
\textbf{Line 1:} \textless{}Observation\textgreater{} a SELF-CONTAINED FULL frame report covering ALL SEVEN labelled fields below, in this order, on EVERY frame (no exceptions, no abbreviations, no same as before, no bare (no change)):\par
\dsfield{description:}{1-2 sentences, holistic summary of THIS frame.}
\dsexample{BAD:}{"A kitchen scene."}
\dsexample{GOOD:}{"Medium shot of a chef in white uniform standing at a stainless-steel counter, chopping carrots with a chef's knife. Steam rising from a pot on the stove in the background."}
\dsfield{actors:}{EVERY distinct person/animal visible. For each: appearance, clothing, role, position. Re-list ALL actors every frame. Include background figures.\par
Always state the TOTAL count (e.g. '2 actors: ...').}
\dsexample{BAD:}{"actors: a person cooking"}
\dsexample{GOOD:}{"actors: 1) chef in white double-breasted jacket, black apron, toque, at center counter --- chopping; 2) sous-chef in blue polo, at back sink --- washing dishes"}
\dsfield{objects:}{salient objects with current STATE and LOCATION. Tag state explicitly.\par
For COUNTABLE objects, report the EXACT number (e.g. '3 cards', '5 bottles') --- never vague ranges like 'approximately 3-5' or 'several'.}
\textbf{Format:} \textless{}object\textgreater{} [state] at \textless{}location\textgreater{}\par
\dsexample{BAD:}{"objects: knife, pot, carrots"}
\dsexample{GOOD:}{"objects: chef's knife [held, in use] in chef's right hand; large stockpot [on stove, steaming] back left; 3 carrots [being chopped] on cutting board center frame"}
\dsfield{actions:}{what each actor is CURRENTLY doing (verb + target + state tag).}
\textbf{Format:} \textless{}actor\textgreater{} \textless{}verb\textgreater{} \textless{}target\textgreater{} [STATE]\par
\textbf{State tags:} [STARTING] / [IN-PROGRESS] / [FINISHED] / [ABORTED]\par
\dsexample{BAD:}{"actions: cooking"}
\dsexample{GOOD:}{"actions: chef chopping carrots [IN-PROGRESS]; sous-chef washing dishes [IN-PROGRESS]"}
For persisting actions, use still ... or continues to ... instead of re-describing.\par
\dsfield{ocr:}{EVERY on-screen text/number VERBATIM. Include language, color, position.}
\dsexample{BAD:}{"ocr: some text on screen"}
\dsexample{GOOD:}{"ocr: 'CHANNEL 5 NEWS' (white text, top-left overlay); scoreboard reads 'HOME 24 - 18 AWAY' (yellow, bottom-right); subtitle: 'What are you doing?' (white, bottom-center)"}
If genuinely no text: "ocr: (none)"\par
\dsfield{scene:}{location + time-of-day + camera framing + indoor/outdoor.}
\dsexample{BAD:}{"scene: indoors"}
\dsexample{GOOD:}{"scene: professional kitchen, daytime (fluorescent lighting), medium-wide shot, eye-level angle, indoor"}
\dsfield{delta:}{what CHANGED from the PREVIOUS frame. This is the MOST IMPORTANT field --- it drives memory compression and retrieval.\par
MUST name at least one concrete, observable change.}
\textbf{Format:} \textless{}subject\textgreater{} \textless{}verb/delta\textgreater{} \textless{}detail\textgreater{}\par
\dsexample{BAD:}{"delta: slight motion" (generic boilerplate)}
\dsexample{BAD:}{"delta: no change" (NEVER allowed)}
\dsexample{BAD:}{"delta: scene continues" (garbage token)}
\dsexample{GOOD:}{"delta: chef's right hand moved 3cm left, knife blade angle changed from 30 degrees to 45 degrees"}
\dsexample{GOOD:}{"delta: steam plume above stockpot expanded upward approximately 5cm; new bubble formation visible"}
\dsexample{GOOD:}{"delta: sous-chef turned head 90 degrees left toward chef; mouth opened (speaking)"}
If you GENUINELY cannot detect any change after active visual inspection, pick a tool (zoom\_crop / track\_query) to find evidence first.\par
ONLY if the frame is bit-identical may you write: "delta: visually identical to t=\{prev\}s --- verified static".\textless{}/Observation\textgreater{}\par
\textbf{CRITICAL when at least two entities of the same kind are visible:} COUNT them exactly (e.g. 5 giants, 3 dogs, 2 cars) and describe their relative spatial order (front/back/left/right/foreground/background, or in a row from left to right: A, B, C style). Identify which one is biggest / closest / leftmost --- this is exactly the kind of fact a future question may rely on.\par
\textbf{Line 2:} \textless{}/Silence\textgreater{} (status tag meaning "observed, no answer yet").\par
Do NOT emit \textless{}/Response\textgreater{}. The system reads \textless{}Observation\textgreater{} into MEMORY and uses \textless{}/Silence\textgreater{} as the round-end marker.\par
\dsheading{Anti-Hallucination Rule}
\dsitem{\textbullet}{Every delta: field MUST reference SPECIFIC VISIBLE elements that YOU CAN SEE in the CURRENT frame(s).}
\dsitem{\textbullet}{Do NOT speculate about off-screen events or future actions.}
\dsitem{\textbullet}{Do NOT infer intentions unless there is explicit visual evidence (e.g., a facial expression, a pointing gesture).}
\dsitem{\textbullet}{If uncertain about a detail (count, color, identity), write [uncertain: \textless{}detail\textgreater{}] rather than guessing.}
\dsitem{\textbullet}{For ANY claim involving a COUNT or COLOR, visually re-examine the frame before writing --- do not trust memory.}
\dsitem{\textbullet}{The delta MUST describe a change you SEE between the two frames; do not invent changes to satisfy the format.}
\dsheading{Initial-frame instruction}
[Round 0 --- initial frame, no previous frame to compare]\par
Write a full observation; set delta to '(initial frame)'.\par
\end{dsprompt}

\par\vspace{2pt}
\begingroup
\captionsetup{font=small,skip=3pt,position=bottom}
\captionof{figure}{Observation-prompt excerpts, including the seven report fields and visual-grounding rules, followed by the separate initial-frame instruction. The literal \texttt{\{prev\}} is retained from the static-frame example.}
\label{fig:prompt-observation}
\label{fig:prompt-grounding}
\endgroup

\subsection{Memory Summarization and Online Recaps}
\label{app:prompt-memory}
The following templates are separate text-only summarization calls; their scheduling and memory links are described in Appendix~\ref{app:memory}. Here, \texttt{\{t0\}} and \texttt{\{t1\}} are interval endpoints, \texttt{\{len(run)\}} is the number of input observations, and \texttt{\{body\}} is their text.

\begin{dsprompt}[left=3mm,right=3mm,
fontupper=\fontencoding{T1}\fontfamily{ptm}\fontsize{10}{12.4}\selectfont,
before upper={\setlength{\parindent}{0pt}\setlength{\parskip}{3.5pt}\raggedright
\renewcommand{\dsheading}[1]{\par\addvspace{7pt}\noindent\textbf{##1}\par\nobreak}}]{Auxiliary memory summarization prompts}
\dsheading{Episodic summary}
\textbf{System message.} You compress a run of streaming-video observations into ONE concise episodic summary. Output 1-3 sentences (at most 120 words total) covering: actors involved, the ordered action chain, final scene state, and any salient changes.\par
Do NOT include per-second timestamps in the body. No markdown, no lists, no preamble --- just the summary text.\par
\textbf{User message.} Window [t=\{t0\}s, t=\{t1\}s], \{len(run)\} events:\par
\{body\}\par
Write the summary now.\par
\dsheading{Long-term record}
\textbf{System message.} You distil a run of streaming-video observations into a durable LONG-TERM FACT record. Prioritise FACT COVERAGE over brevity: capture EVERY concrete fact present --- all actors (with exact counts), all key objects (with counts), the action(s) performed and their final state (started/in-progress/finished), locations and spatial relations, and any visible on-screen text/numbers. Prefer listing more facts to omitting them. Do NOT speculate beyond the observations and do NOT invent details.\par
Output plain prose or semicolon-separated facts (no markdown, no preamble, up to approximately 6 sentences).\par
\textbf{User message.} Segment [t=\{t0\}s, t=\{t1\}s], \{len(run)\} observations:\par
\{body\}\par
Write the fact-covering long-term record now.\par
\end{dsprompt}

\par\vspace{2pt}
\begingroup
\captionsetup{font=small,skip=3pt,position=bottom}
\captionof{figure}{Complete system and user templates for the two auxiliary memory-summarization calls.}
\label{fig:prompt-memory}
\endgroup

The Recap instruction applies at semantic boundaries within frame-processing turns before or after query arrival. It does not require a separate summarization call.

\begin{dsprompt}[]{Online Recap instruction}
\dsheading{Recap on semantic boundaries (REQUIRED when applicable)}
If THIS frame marks a semantic boundary --- an event you were narrating just FINISHED, a SCENE CHANGE happened, an ACTOR ENTERED / LEFT / visibly CHANGED, or a KEY OBJECT appeared/disappeared/changed state --- additionally emit a \textless{}Recap t0=\allowbreak{}\textless{}sec\textgreater{} t1=\allowbreak{}\textless{}sec\textgreater{} kind=\allowbreak{}\textless{}event\_end|\allowbreak{}scene\_change|\allowbreak{}actor\_enter|\allowbreak{}actor\_exit|\allowbreak{}actor\_change|\allowbreak{}object\_change|\allowbreak{}milestone\textgreater{} state=\allowbreak{}\textless{}finished|\allowbreak{}started|\allowbreak{}in\_progress|\allowbreak{}aborted\textgreater{}\textgreater{} ... \textless{}/Recap\textgreater{} block in this same turn (BEFORE \textless{}/Silence\textgreater{}).\par
The state=\allowbreak{} attribute is REQUIRED on event\_end / milestone / object\_change / actor\_change recaps. It tells future readers (and event-state/answerability meta-questions) the EXACT progression status of the event:\par
\dsitem{\textbullet}{state=\allowbreak{}finished: the action / step is FULLY COMPLETED (use this when narrating event\_end). Future event-state queries can use this as completed evidence.}
\dsitem{\textbullet}{state=\allowbreak{}started: the action has just begun, evidence is now observable but not yet completed. Future answerability queries can use this as onset evidence.}
\dsitem{\textbullet}{state=\allowbreak{}in\_progress: the action is mid-stream; completion not yet reached.}
\dsitem{\textbullet}{state=\allowbreak{}aborted: the actor visibly stopped or abandoned the action without finishing.}
The Recap body MUST start with an explicit verb phrase that names the state, e.g. "FINISHED: \textless{}action\textgreater{} at t=\allowbreak{}12s", "STARTED: \textless{}action\textgreater{} at t=\allowbreak{}8s, currently in progress". This single phrase is what future memory\_search will surface for state queries.\par
The Recap should be 3--8 sentences and MUST cover: every actor involved (with appearance), the full ordered action chain in [t0,t1], \textbf{final state of key objects (use one of: present/absent, open/closed, on/off, full/empty, moving/static)}, spatial layout at the boundary, any on-screen text / dialogue verbatim, and why this is a boundary. The system pins it as durable long-term memory.\par
Do NOT emit Recap on quiet frames; do NOT emit more than one every approximately 5 seconds.\par
\end{dsprompt}

\par\vspace{2pt}
\begingroup
\captionsetup{font=small,skip=3pt,position=bottom}
\captionof{figure}{Instruction excerpt for Recap blocks emitted by the streaming agent at semantic boundaries.}
\label{fig:prompt-recap}
\endgroup

\subsection{Query-Driven Planning}
\label{app:prompt-planner}
The complete planning-phase instruction is followed by its acknowledgement and the template inserted into subsequent answering calls.

\begin{dsprompt}[]{Planning prompt}
[Phase] QUESTION POSED --- PLANNING STEP (analyse BEFORE you answer). The question is below (and pinned to MEMORY). Do NOT answer yet and do NOT call tools this turn. First THINK about what the question truly asks, then emit ONE \textless{}Plan\textgreater{} block:\par
\dscode{\textless{}Plan\textgreater{}}
\dsplanfield{need:}{\textless{}in your own words, the exact thing being asked --- the TARGET and its ANCHOR (the moment/event/actor it pivots on)\textgreater{}}
\dsplanfield{pov:}{\textless{}FIRST-PERSON CHECK --- does the question say "I"/"me"/"my"? If so, "I" am the CAMERA WEARER, NOT any person fully visible in the frame.\par
The camera wearer's own hands/arms/legs appear entering from the BOTTOM or EDGE of the frame; a person sitting across the table fully facing the camera is the OTHER person, NOT me. \par State clearly WHO "I" am and WHO the other people are here.\par
Note my own body is often only partially visible or off-screen, so a question about what I do may not be answerable by looking at me directly\textgreater{}}
\dsplanfield{info\_type:}{\textless{}what KIND of evidence settles it --- e.g. temporal order of actions / spatial direction or position / existence (is X even present?) / exact count / object identity / attribute like color\textgreater{}}
\dsplanfield{evidence\_locus:}{\textless{}work out WHERE the deciding evidence for THIS question lives --- in your own words, no default direction. Weigh three possibilities even-handedly and pick whichever the question truly calls for:\par
(i) it is in the CURRENT frame / the moment the question is asked (what is happening or visible right now);\par
(ii) it is in something that has ALREADY happened (recorded in MEMORY / a specific past moment to re-look);\par
(iii) it has NOT happened yet and you must keep observing until it does.\par
Do not assume it is in the past just because MEMORY is handy, nor in the present just because a frame is in front of you.\par
Check the ANCHOR the question pivots on (e.g. the event in "what happens AFTER X" / "before X"): has that anchor --- and the thing asked relative to it --- actually appeared yet? If not, the evidence is not in hand and answering now would just guess from whatever is nearest. \par Then say how you'll obtain it: look at the current frame, read/search a MEMORY layer (for what terms), replay a specific moment (which t and why), or keep watching. \par If the target is hard to observe directly (occluded, off-screen, or your own body), infer it INDIRECTLY from observable traces --- how the objects it acts on changed, the before/after scene difference, or the effect it left\textgreater{}}
\dsplanfield{answerability:}{\textless{}look at the ACTUAL listed OPTIONS (A/B/C/D...) as a CLOSED set --- your final answer MUST be one of them verbatim.\par
Check whether one of THOSE listed options literally says "Unable to answer" / "cannot determine" (or similar). Do NOT count the generic instruction to "give a best-guess if uncertain" as such an option --- that is NOT a choice, and there is no "implied" unable option.\par
If such a choice IS listed: your first job is to VERIFY the target exists before locating it, and you may pick it when the evidence truly shows the target is absent. \par If NO such choice is listed: refusing is FORBIDDEN --- even with weak evidence you must commit the single MOST LIKELY listed option, never "Unable to answer". State here which case this is\textgreater{}}
\dsplanfield{meta:}{\textless{}is the question asking whether something CAN be answered / WILL happen / has STARTED-vs-FINISHED (a meta/state question), rather than a plain fact? yes/no + why\textgreater{}}
\dsplanfield{strategy:}{\textless{}turn your evidence\_locus reasoning into a concrete plan: given WHERE you decided the evidence lives, spell out which memory layer to read/search (and for what terms), whether the current frame suffices, and which exact t to replay if you need to confirm.\par
E.g. for an action-order question: locate the exact t of the anchor action, then read the NEXT action change after it.\par
For a direction question: compare two frames a few seconds apart, do NOT track.\par
For an existence question: search once or twice; if empty AND "Unable to answer" is a LISTED option, answer it --- otherwise commit the most likely listed option. \par IMPORTANT --- if info\_type is a CURRENT-FRAME percept (what the person is doing RIGHT NOW, an on-screen object/text/attribute, an immediate action): the answer is in the CURRENT frame --- LOOK at it and answer directly.\par
Do NOT replay the current moment (it just re-shows the same frame), do NOT browse other frames, and use at most ONE zoom\_crop if a detail is unclear.\par
Commit quickly --- over-using tools on a see-it-now question adds noise and flips close options (e.g. connecting vs disconnecting).\textgreater{}}
\dscode{\textless{}/Plan\textgreater{}}
End this turn with \textless{}/Silence\textgreater{} right after the \textless{}/Plan\textgreater{} block (you will be asked again immediately to act on this plan).\par
\end{dsprompt}

The acknowledgement confirms that the plan was parsed. The stored plan is supplied separately in later calls.

\begin{dsprompt}[]{Runtime acknowledgement}
\dscode{\textless{}tool\_result\textgreater{}\{"name": "\_plan", "ok": true, "data": \{"note": "Plan recorded. Now ACT on it: gather the evidence your strategy named, then answer. It will be shown to you each turn."\}\}\textless{}/tool\_result\textgreater{}}
\end{dsprompt}

\par\vspace{2pt}
\begingroup
\captionsetup{font=small,skip=3pt,position=bottom}
\captionof{figure}{Complete planning-phase instruction with all seven fields, followed by a runtime acknowledgement of the parsed plan.}
\label{fig:prompt-planner}
\endgroup

In the runtime template below, angle-bracketed values are editorial placeholders. Nonempty plan fields contain the model's analysis; the contract's \texttt{evidence\_locus} is a rule-derived category. The template illustrates the format, not a recorded execution.

\begin{dsprompt}[]{Persistent plan and evidence guidance (runtime template)}
[PLAN] Your own analysis of this question --- FOLLOW IT:\par
\dsitem{-}{need: \textless{}stored need\textgreater{}}
\dsitem{-}{pov: \textless{}stored pov\textgreater{}}
\dsitem{-}{info\_type: \textless{}stored info\_type\textgreater{}}
\dsitem{-}{evidence\_locus: \textless{}stored evidence\_locus\textgreater{}}
\dsitem{-}{answerability: \textless{}stored answerability\textgreater{}}
\dsitem{-}{meta: \textless{}stored meta\textgreater{}}
\dsitem{-}{strategy: \textless{}stored strategy\textgreater{}}
\dsheading{[EVIDENCE CONTRACT --- compiled from your plan]}
\dsitem{-}{retrieval\_query: \textless{}question and planned-need keywords\textgreater{}}
\dsitem{-}{evidence\_locus: \textless{}current, future, past, or mixed\textgreater{}}
\dsitem{-}{tool\_policy: \textless{}guidance selected from the temporal category and evidence type\textgreater{}}
\dsitem{-}{decision\_rule: answer only from evidence matching the target and anchor; options are output choices, not retrieval evidence.}
\end{dsprompt}

\par\vspace{2pt}
\begingroup
\captionsetup{font=small,skip=3pt,position=bottom}
\captionof{figure}{Runtime template for persistent plan fields and the derived evidence contract. Placeholder values identify dynamic content rather than a recorded execution.}
\label{fig:prompt-plan-persistence}
\endgroup

\subsection{Evidence Gathering and Answering}
\label{app:prompt-evidence}
These excerpts omit task-specific inputs and additional runtime reminders.

For fixed-choice questions, the \texttt{abstain} instructions permit an ``Unable to answer'' response only when such an option is explicitly listed. Otherwise, the agent must select a listed choice, as specified by the planner's \texttt{answerability} field.

\begin{dsprompt}[]{Tool use and question arrival}
\dsheading{Tool-call protocol}
Whenever you want to call a tool, emit \emph{exactly} the following block as part of your response (you may emit multiple such blocks):\par
\dscode{\textless{}tool\_call\textgreater{}\{"name": "\textless{}tool\_name\textgreater{}", "arguments": \{...\}\}\textless{}/tool\_call\textgreater{}}
After the tool block(s), STOP. The system will run the tools and feed \textless{}tool\_result\textgreater{}\{...\}\textless{}/tool\_result\textgreater{} back to you. Only after you have enough evidence, finish with \textless{}/Response\textgreater{} \textless{}your final answer\textgreater{}.\par
\dsheading{Question-arrival instruction}
[Phase] QUESTION POSED. The question below is being asked NOW (and has been pinned to MEMORY).\par
\dsheading{Read-the-question protocol (BEFORE deciding)}
Silently parse the question text into:\par
\dsitem{\textbullet}{TARGET: what is being asked (an entity, an action, a count, a Yes/No truth value, an on-screen text string, ...).}
\dsitem{\textbullet}{ANCHOR: the time / event / actor it pivots on (e.g. right after loading the wheel, the person we communicated with while chopping, under the monkey at this moment).}
\dsitem{\textbullet}{FORMAT: option letter, Yes/No, integer 0..N, free-form span --- match the wording exactly.}
Then check MEMORY for events that match the ANCHOR; the TARGET is whatever is reported in those events.\par
If MEMORY only contains the ANCHOR but not the TARGET, the answer probably lies a few frames later --- emit \textless{}/Silence\textgreater{} and wait, OR replay to a likely frame.\par
Decide whether you can answer with the evidence accumulated so far (see MEMORY) or need to watch more frames (emit \textless{}/Silence\textgreater{} to wait).\par
GROUND YOUR ANSWER IN PIXELS WHEN IT HELPS: if the answer hinges on a SPECIFIC PAST moment (an earlier action / object state / who was present), replay(t=\allowbreak{}...) that moment and verify with your own eyes before committing.\par
YOU decide whether a replay is needed and how many: questions about the CURRENT frame usually need none; questions about the past may need one or a few. Do not over-replay when the current frame already answers it.\par
BEFORE answering, prefer a quick tool call if it would resolve uncertainty:\par
\dsitem{\textbullet}{To re-examine an earlier moment, use replay.}
\dsitem{\textbullet}{To re-read your notes, use memory\_read or memory\_list.}
\dsitem{\textbullet}{To inspect fine detail, use zoom\_crop.}
\dsitem{\textbullet}{If there is genuinely no visual evidence at all, abstain.}
One or two well-chosen tool calls almost always beat a blind guess.\par
\end{dsprompt}

\begin{dsprompt}[]{Continued observation and evidence gathering}
[Phase] POST-QUESTION. The question is pinned in MEMORY (look for the [Q] ... entry). Continue watching the new frame.\par
\dsheading{Read-the-question protocol (every round)}
RE-READ the pinned [Q] ... line and parse it into TARGET / ANCHOR / FORMAT (see QUESTION POSED phase). Then ask: does the latest frame OR any new MEMORY event satisfy the ANCHOR? If yes, the TARGET is the fact reported there.\par
\textbf{First-Person POV:} when the camera shows bare feet/legs/torso at the bottom of the frame, the camera wearer is 'I'. A question asking 'who did I communicate TO' or 'who did I interact WITH' is asking about the OTHER person --- NOT the one visible through the first-person perspective (that is 'I').\par
Trace both people's timelines in MEMORY: the camera-wearer ('I') appears as partial body at frame bottom; the interaction partner is a fully-visible actor.\par
If the answer depends on a SPECIFIC PAST frame, replay(t=\allowbreak{}...) that moment to CONFIRM with your eyes first; if the CURRENT frame or MEMORY already answers it clearly, just emit \textless{}/Response\textgreater{} \textless{}answer\textgreater{}. Otherwise emit \textless{}/Silence\textgreater{}.\par
Use replay WHEN a past moment needs verifying, but YOU decide if it is needed and how many; don't replay when the current frame already settles the answer.\par
If something in MEMORY is fuzzy or this frame hints at the answer, use a tool BEFORE answering --- replay to revisit a past moment, zoom\_crop for detail, memory\_read to recall a pinned fact, asr\_segment for audio, or abstain if there is genuinely no evidence at all.\par
\textbf{CRITICAL:} After EVERY tool result, FIRST write a 1-2 line observation of what the tool revealed (especially: counts, text, identities, state changes) BEFORE you decide to call another tool.\par
Do NOT chain tools blindly --- each tool call must be justified by a specific gap in what you have ALREADY observed.\par
If the tool result confirms what you need, COMMIT with \textless{}/Response\textgreater{} NOW instead of calling more tools. A zoom\_crop that returns a clear image of the target region IS ENOUGH --- write your observation and answer.\par
If THIS frame marks a semantic boundary (event ended / scene change / actor enter-exit-change / key object change), ALSO emit a \textless{}Recap t0=\allowbreak{}\textless{}sec\textgreater{} t1=\allowbreak{}\textless{}sec\textgreater{} kind=\allowbreak{}\textless{}...\textgreater{}\textgreater{} ... \textless{}/Recap\textgreater{} block (3--8 sentences, see OBSERVE-ONLY phase instructions for Recap format) --- the system pins it as durable memory.\par
\dsheading{How to use the [RETRIEVED] block}
The [RETRIEVED ...] block above is the system's best-guess evidence for your question. HOW TO USE IT:\par
\dsitem{\textbullet}{Each path\#N shows a chain through memory layers (from wk to ep to lt). Follow the chain --- the working-layer event is closest to the original pixels.}
\dsitem{\textbullet}{Cross-check: if a fact appears consistently across working + episodic, it is reliable. If only in long\_term, verify via replay.}
\dsitem{\textbullet}{It is NOT exhaustive --- use memory\_search with different keywords if needed.}
\dsheading{Tool selection (use BEFORE each call)}
\dsitem{1.}{For a MOVING subject across frames, use track\_init (NOT zoom\_crop).}
\dsitem{2.}{If the answer depends on a PAST frame, use memory\_search first; if detail is missing, use replay.}
\dsitem{3.}{If the cue is likely AUDIO, use asr\_segment.}
\dsitem{4.}{For STATIC fine detail in the CURRENT frame (text, label, logo), use zoom\_crop.}
\dsitem{5.}{To remember a durable fact, use memory\_write(pin=\allowbreak{}true).}
Anti-pattern: after 2 or more zoom\_crops with no new fact, switch tools.\par
\end{dsprompt}

\par\vspace{2pt}
\begingroup
\captionsetup{font=small,skip=3pt,position=bottom}
\captionof{figure}{Excerpts of the shared tool-call protocol and post-query instructions for evidence gathering, visual verification, and answering.}
\label{fig:prompt-tools}
\label{fig:prompt-query-arrival}
\label{fig:prompt-continued}
\endgroup

\endgroup

\clearpage
\section{Additional Qualitative Examples}
\label{app:qualitative}


\begingroup
\setlength{\intextsep}{5pt}
\captionsetup{font=small,skip=4pt}
\begin{figure}[H]
\centering
\includegraphics[width=\textwidth]{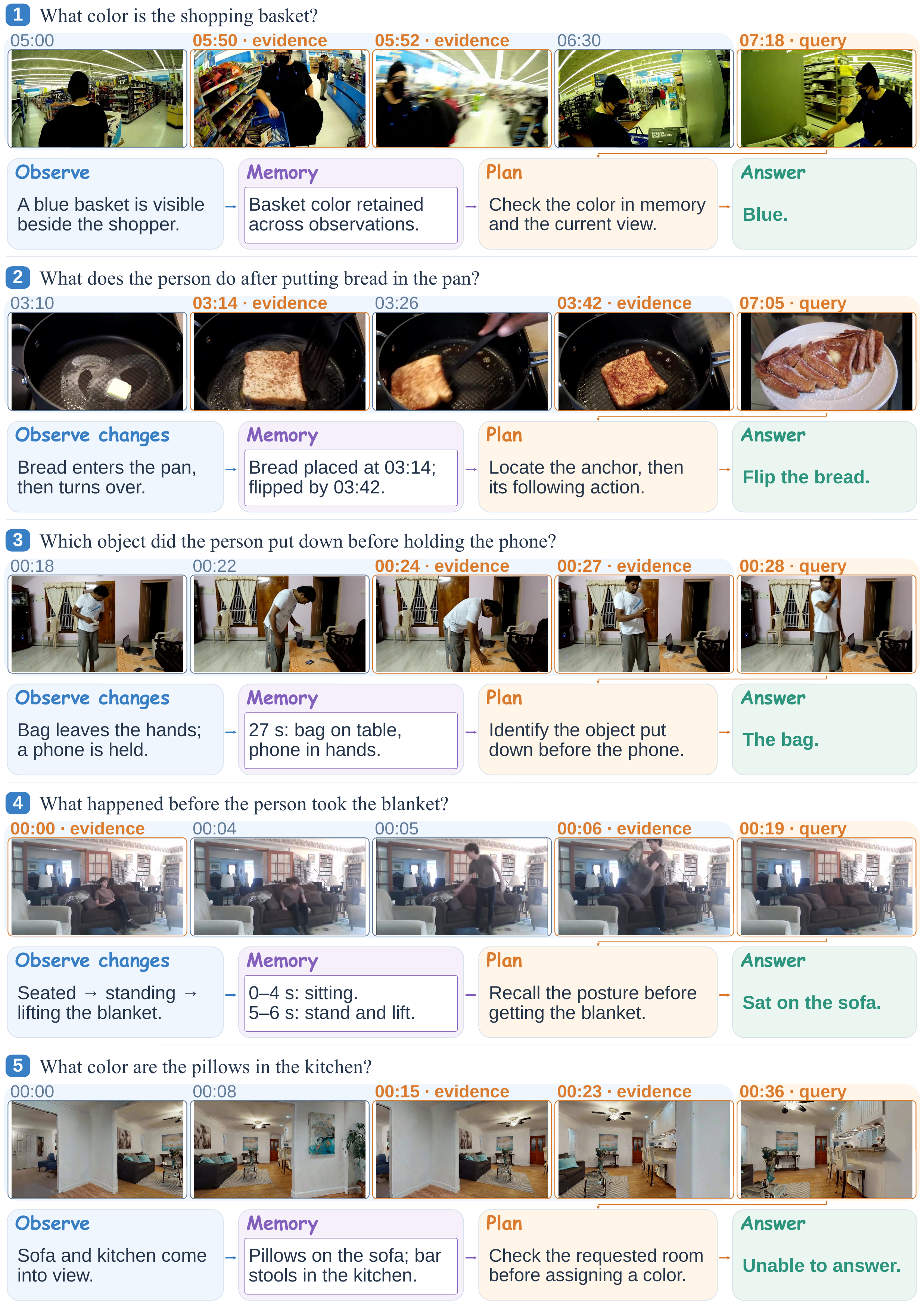}
\caption{OVO-Bench examples of temporal reasoning, attribute recall, and checking query premises.}
\label{fig:appendix-cases-ovo}
\end{figure}

\clearpage
\begin{figure}[H]
\centering
\includegraphics[width=\textwidth]{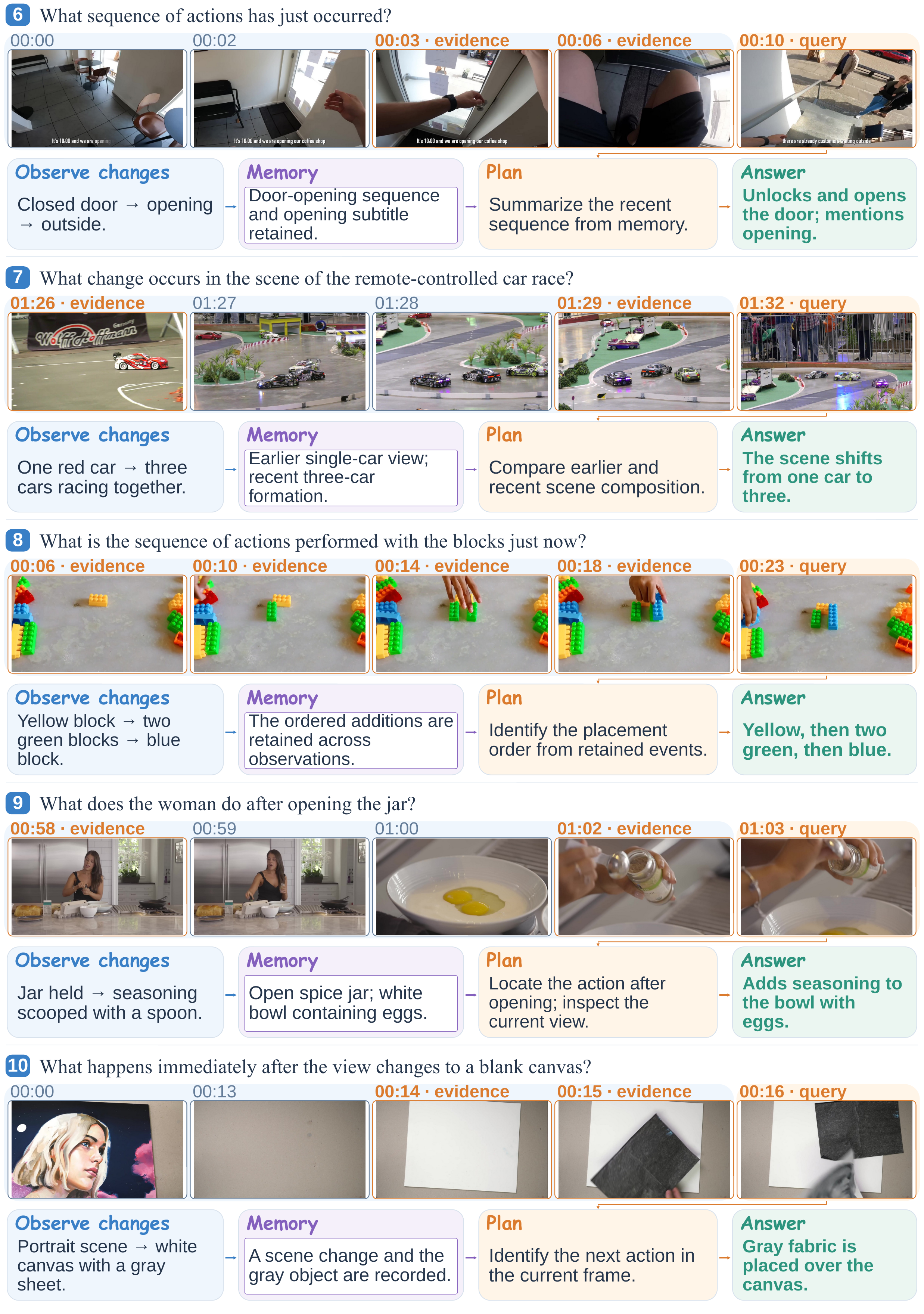}
\caption{StreamingBench examples of action sequencing, scene changes, and visual reasoning.}
\label{fig:appendix-cases-sb}
\end{figure}

\clearpage
\begin{figure}[H]
\centering
\includegraphics[width=\textwidth]{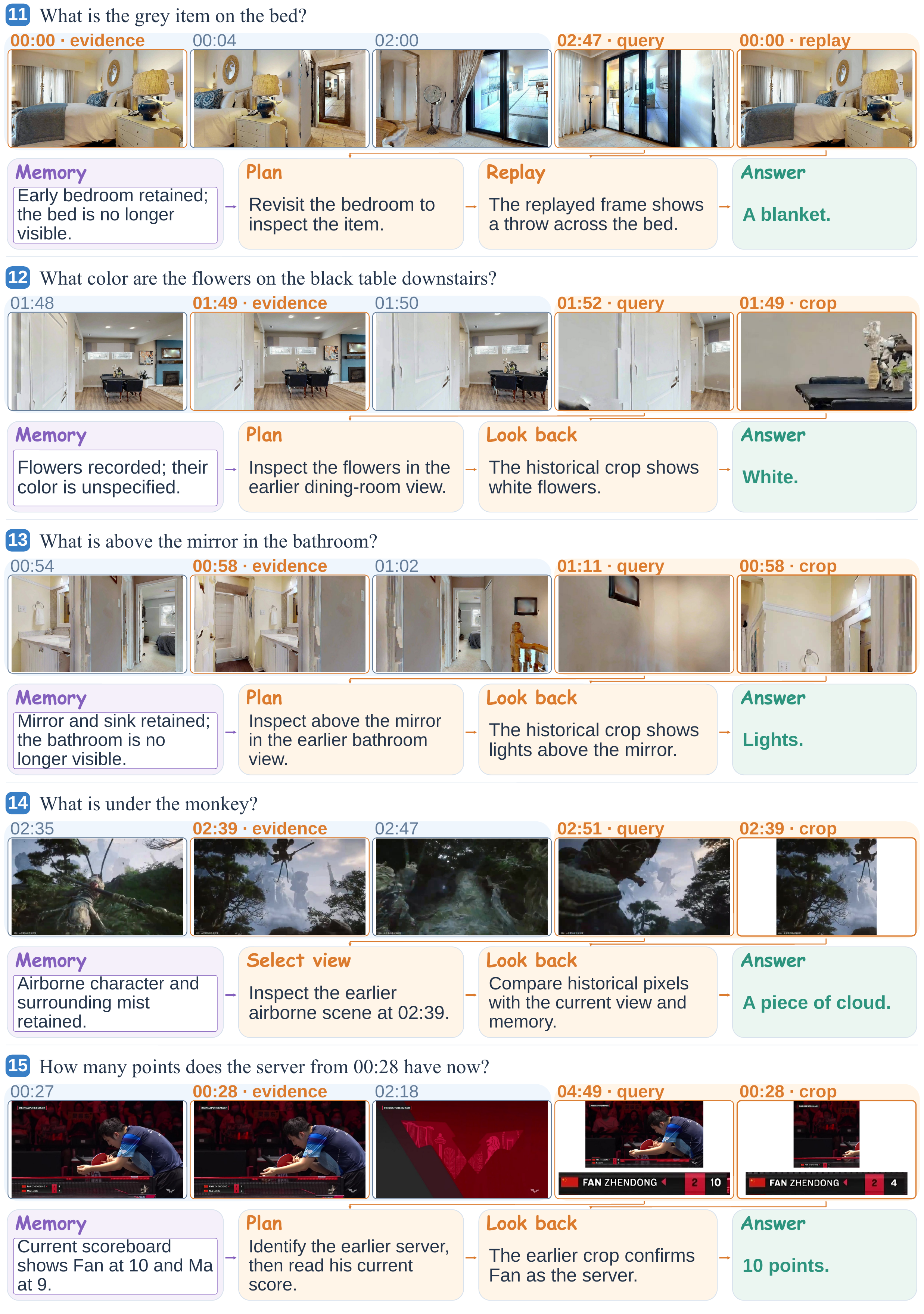}
\caption{Some examples of using replay or historical crops.}
\label{fig:appendix-cases-replay}
\end{figure}
\endgroup

\end{document}